# A survey of synthetic data augmentation methods in computer vision

Alhassan Mumuni[1]*, Fuseini Mumuni[2] and Nana Kobina Gerrar[1]

**Abstract**—The standard approach to tackling computer vision problems is to train deep convolutional neural network (CNN) models using large-scale image datasets which are representative of the target task. However, in many scenarios, it is often challenging to obtain sufficient image data for the target task. Data augmentation is a way to mitigate this challenge. A common practice is to explicitly transform existing images in desired ways so as to create the required volume and variability of training data necessary to achieve good generalization performance. In situations where data for the target domain is not accessible, a viable workaround is to synthesize training data from scratch—i.e., synthetic data augmentation. This paper presents an extensive review of synthetic data augmentation techniques. It covers data synthesis approaches based on realistic 3D graphics modeling, neural style transfer (NST), differential neural rendering, and generative artificial intelligence (AI) techniques such as generative adversarial networks (GANs) and variational autoencoders (VAEs). For each of these classes of methods, we focus on the important data generation and augmentation techniques, general scope of application and specific use-cases, as well as existing limitations and possible workarounds. Additionally, we provide a summary of common synthetic datasets for training computer vision models, highlighting the main features, application domains and supported tasks. Finally, we discuss the effectiveness of synthetic data augmentation methods. Since this is the first paper to explore synthetic data augmentation methods in great detail, we are hoping to equip readers with the necessary background information and in-depth knowledge of existing methods and their attendant issues.

## 1 INTRODUCTION

### 1.1 Background

Currently, deep learning is the most important technique for solving many complex machine vision problems. State-of-the-art deep learning models typically contain very large number of parameters that need to be learned in order to characterize a wide range of visual phenomena. Moreover, as a result of the enormous appearance variations of real-world objects and scenes, there is often the need to introduce various variations of the available data during training. Consequently, training deep learning models require very huge amount of annotated data to guarantee good generalization performance and avoid overfitting. However, data collection and annotation are often time-consuming and costly exercises. Instead of attempting to collect very large quantities of annotated data, it is often more practical to create new samples artificially. Data augmentation is the process of creating new data to artificially extend training set. Typically, the process of augmentation involves performing transformations on the original data so as to alter them in a particular way. The transformation operations usually change the visual characteristics of the data but preserve their labels. Data augmentation (DA) can, thus, be seen as a means for simulating real-world behavior such as the visual appearance of objects and scenes under different view angles, pose variations, object deformations, lens distortions and other camera artifacts. In practice, there are several situations in which the training of machine learning (ML) models may require data augmentation. The commonest scenarios include the following:

- Training a deep learning models but the quantity of training data is small
- Adequate training data exists but is of perceptually poor quality (e.g., low resolution, hazy or blurry)
- Available training data is not representative of the target data (e.g., does not have adequate appearance variations)
- The proportion of various classes is skewed (imbalanced data)
- Data is only available for one condition (e.g., bright day) but there is the need to train models to perform inference under different set of conditions (e.g., night, rainy or foggy weather)
- There is no practical way to access data for training (e.g., excessive cost or restriction)

The first four problems can be adequately solved by manipulating the existing data to produce additional data that enhances the overall performance of the trained model. In the case of the last two problems, however, the only viable solution is to create new training data.

### 1.2 Significance of synthetic data augmentationn

As discussed earlier, the commonest approach to data augmentation is to transform training data in various ways. However, in application scenarios where no training data exists naturally, or where its collection is too costly, it often becomes impractical to create additional training data using

. [1]*Cape Coast Technical University, Cape Coast, Ghana.*
*\*Corresponding author, E-mail: alhassan.mumuni@cctu.edu.gh*
- [2]*University of Mines and Technology, UMaT, Tarkwa, Ghana*

the aforementioned methods. Moreover, many computer vision tasks are often use-case sensitive, requiring task-specific data formats and annotation schemes. This makes it difficult for broadly-annotated, publicly-available large-scale datasets to meet the specific requirements of these tasks. In these cases, the only viable approach is to generate training data from scratch. Modern image synthesis methods can simulate different kinds of task-specific, real-world variability in the synthesized data. They are particularly useful in applications such as autonomous driving and navigation [1], [2], pose estimation [3], [4], affordance learning [5], [6], object grasping [7], [8] and manipulation [9], [10], where obtaining camera-based images is time-consuming and expensive. Moreover, in some applications, bitmap pixel images may simply be unsuitable. Data synthesis methods can readily support non-standard image modalities such as point clouds and voxels. Approaches based on 3D modeling also provide more scalable resolutions as well as flexible content and labeling schemes that is adapted for the specific use-case.

### 1.3 Motivation for this survey

Data augmentation approaches based on data synthesis are becoming increasingly important in the wake of severe data scarcity in many machine learning domains. In addition, the requirements of emerging machine vision applications such as autonomous driving, robotics and virtual reality are increasingly becoming difficult to be met using traditional data transformation-based augmentation. For this reason, data synthesis has become an important means to provide quality training data for machine learning applications. Unfortunately, however, while many surveys on data augmentation approaches exist, very few works deal with synthetic data augmentation methods. This work is motivated by the lack of adequate discussion on this important class of techniques in the scientific literature. Consequently, we aim to provide an in-depth treatment of synthetic data augmentation methods to enriched the current literature on data augmentation. We discuss the various issues on data synthesis in detail, including concise information about the main principles, use-cases and limitations of the various approaches.

### 1.4 Outline of work

In this work, we first provide a broad overview of data augmentation in Section 2 and provide a concise taxonomy of synthetic data augmentation approaches in Section 3. Further, in Section 4 through 7, we explore in detail the various techniques for synthesizing data for machine vision tasks. Here we discuss the important principles, approaches, use-cases and limitations of each of the main classes of methods. The approaches surveyed in this work are generative modeling, computer graphics modeling, neural rendering, and neural style transfer (NST). We present a detailed discussion of each of these approaches in the following sections. We also compare the advantages and disadvantages of these classes of data synthesis methods. We summarize the main features of common synthetic datasets in Section 8. In Section 9 we discuss the effectiveness of synthetic data augmentation in machine vision domains. We present a summary of the main issues in Section 10 and outline promising directions for future research in Section 11. Finally, conclude in Section 12. A detailed outline of this survey is presented in Figure 1.

## 2 OVERVIEW OF DATA AUGMENTATION METHODS

Geometric data augmentation methods such as affine transformations [11], projective transformation [12] and nonlinear deformation [13] are aimed at creating various transformations of the original images to encode invariance to spatial variations resulting from, for example, changes in object size, orientation or view angles. Common geometric transformations include rotation, shearing, scaling or resizing, nonlinear deformation, cropping and flipping. On the other hand, photometric techniques – for example, color jittering [14], lighting perturbation [15], [16] and image denoising [17] – manipulate the qualitative properties – for example, image contrast, brightness, color, hue, saturation and noise levels – and thereby render the resulting deep learning models invariant to changes in these properties. In general, to ensure good generalization performance in different scenarios, it is often necessary to apply many of these procedures simultaneously.

Recently, more advanced data augmentation methods have become common. One of the most important class of techniques [18], [19], [20], [21] is based on transforming different image regions discretely instead of uniformly manipulating the entire input space. This type of augmentation methods have been shown to be effective in simulating complex visual effects such as non-uniform noise, non-uniform illumination, partial occlusion and out-of-plane rotations.

The second main direction of data augmentation exploits feature space transformation as a means of introducing variability of training data. These regularization approaches manipulate learned feature representations within deep CNN layers to transform the visual appearance of the underlying images. Examples of feature-level transformation approaches include feature mixing [18], [22], feature interpolation [23], feature dropping [24] and selective augmentation of useful features [25]. These methods do not lead to semantically meaningful alterations. Nonetheless, they have proven very useful in enhancing the performance of deep learning models. The third direction is associated with the automation of the augmentation process. To achieve this, typically, different transformation operations based on traditional image processing techniques are applied to manually generate various primitive augmentations. Optimization algorithms are then used to automatically find the best model hyperparameters, as well as the augmentation types and their magnitudes for the given task.

The approaches described above are realizable only when training data exists, and the goal of augmentation is to transform the available data to obtain desirable features. This work focuses on approaches that seek to generate novel training data even in cases where data for the target task is inaccessible.

Several survey works (e.g., [26], [27], [28], [28], [29], [30]) have explored data augmentation in great detail. Shorten et al. [27], in particular, present a broad discussion of important data augmentation methods. However, like most previous surveys, their coverage of data synthesis methods is rather limited.

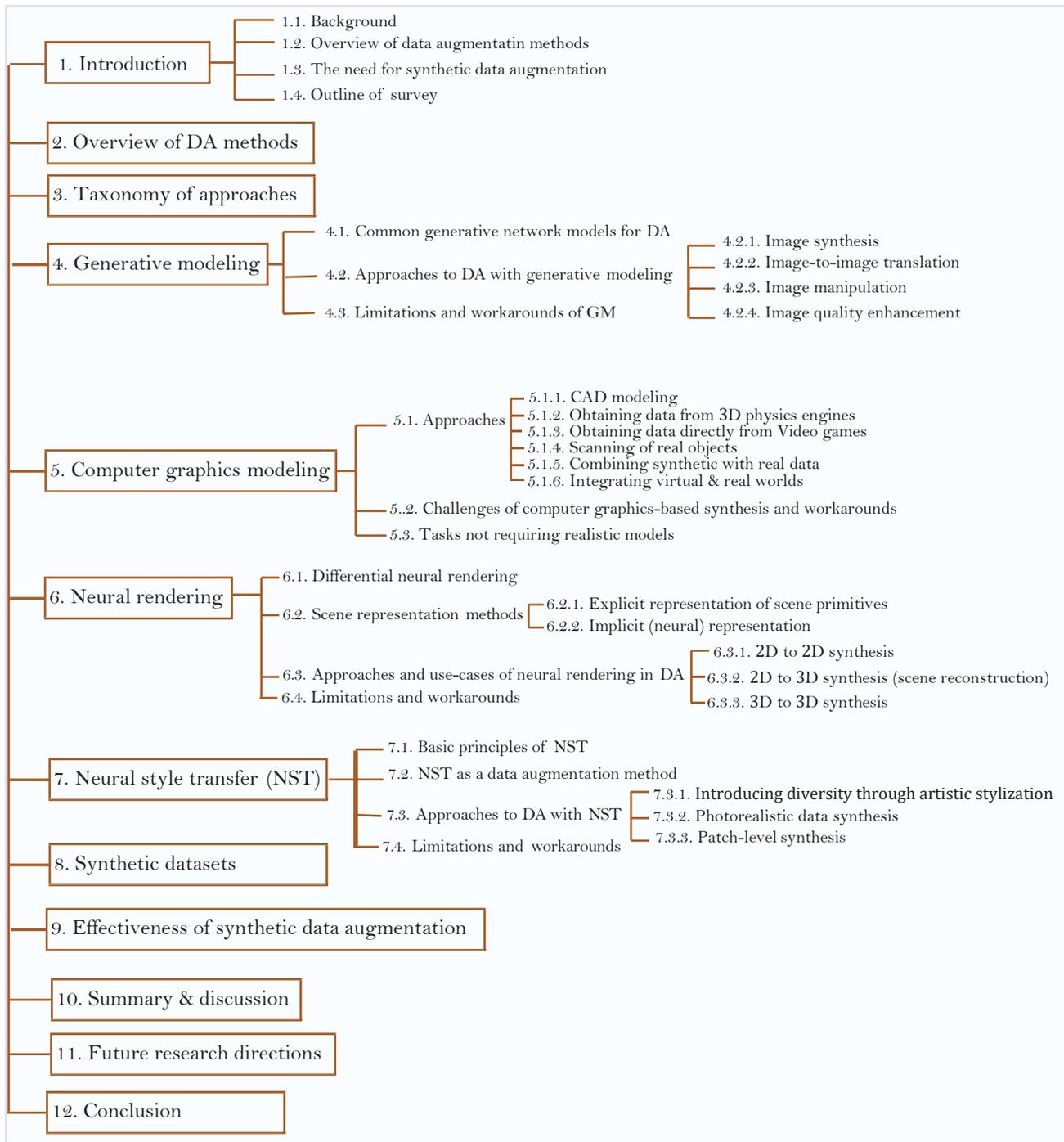

Figure 1. Detailed outline of work

To address this gap, in this surveys, we focus mainly on data augmentation techniques that generate synthetic data for training machine learning models in computer vision domains. data augmentation methods. The main approaches covered here are methods based on generative AI, procedural data generation using 3D CAD tools and game engines, differential neural rendering and neural style transfer. We consider that such a narrow scope will enable us to provide a much detailed treatment of topic and its important issues while at the same time maintain a relatively concise volume.

## 3 TAXONOMY OF SYNTHETIC DATA AUGMENTATION METHODS

In practice, four main classes of synthetic data generation techniques are commonly used:

- generative modeling
- computer graphics modeling
- neural rendering
- neural style transfer

Generative modeling methods rely on learning the inherent statistical distribution of input data in order to (automatically) generate new data. The second class of approaches, computer graphics modeling, are based on the elaborate process of manually constructing 3D models of objects and scenes with the aid of computer graphics tools. Neural rendering approaches are based on special methods for constructing novel data using conventional feedforward neural network architectures are described. They allow deep neural networks to generate completely new images by learning intermediate 3D representations. Where needed, the generated 3D data can be used for training. The fourth class of methods are known as neural style transfer. These approaches combine features of different semantic levels extracted from different images to create a new set of images. A general classification of synthetic data augmentation approaches is depicted in Figure 2.

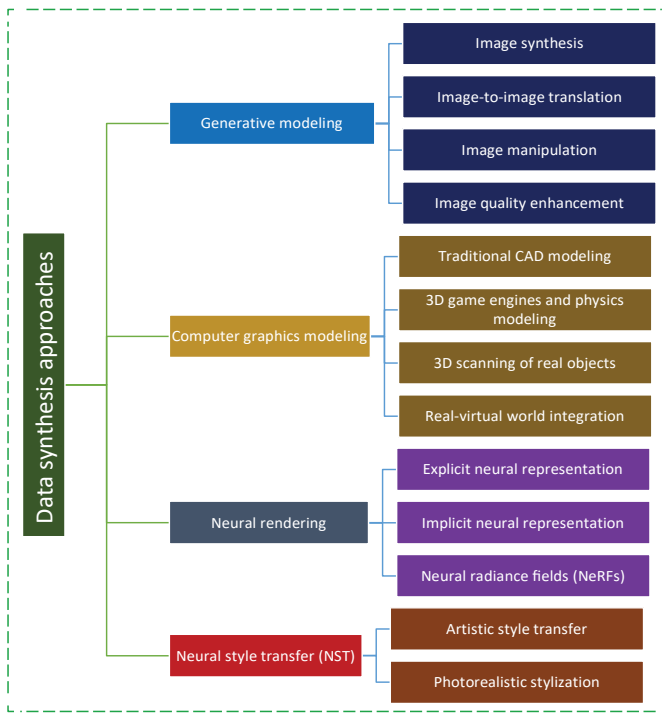

Figure 2. Taxonomy of synthetic data augmentation approaches

## 4 GENERATIVE MODELING

Generative AI techniques present the most promising prospect for generating synthetic datasets for complex computer vision tasks. Generative modeling methods are a class of deep learning techniques that utilize special deep neural network architectures to learn holistic representation of the underlying categories in order to generate useful synthetic data for training deep learning models. Generally, they work by learning possible statistical distributions of the target data using noise or examples of target data as input. This knowledge about the distribution of training data, thus, can enable them to generate complex representations. Examples of generative models include Boltzmann machines (BMs) [31] and restricted Boltzmann machines (RBMs) [32], generative adversarial networks (GANs) [33], variational autoencoders (VAEs) [34], autoregressive models [35] and deep belief networks (DBNs) [36]. Currently, GANs and VAEs and their various variants such as [37], [38], [39] are the most popular neural network architectures for generative modeling.

### 4.1 Common generative AI models for data generation

#### a. Generative adversarial network (GAN)

The general structure of the GAN is shown in Figure 3a. For image generation, the basic working principle of the GAN is as follows. A generator samples multi-dimensional noise from a random distribution and converts this noise into a representation similar to real images. A discriminator then tries to distinguish between real images and the artificially generated samples and provides feedback about its predictions to the generator. The generator, aiming to produce samples that are indistinguishable from real ones, iteratively refines its prediction based on the feedback error computed by the loss function. In a similar way, the discriminator uses its own loss to improve subsequent predictions. This process eventually leads to the model generating high-quality images. The vanilla GAN uses fully connected multi-layer deep neural network architecture for the implementation of both Generator and Discriminator sub-models.

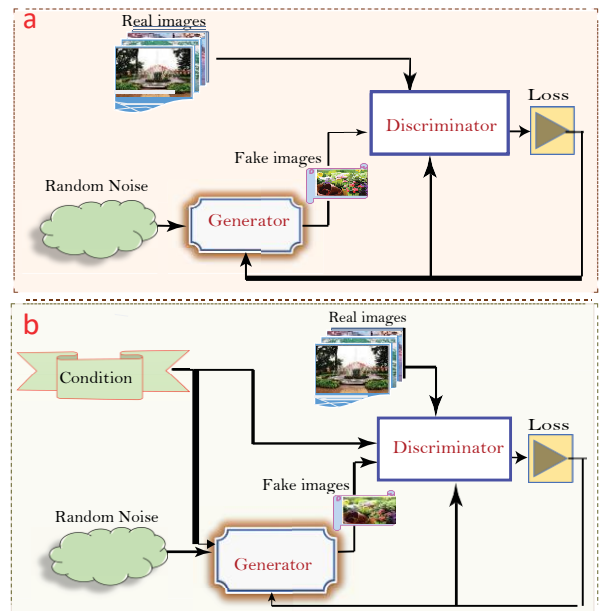

Figure 3. Functional block diagram of a basic GAN (a) and Conditional GAN (b) models.

In the Conditional GAN (cGAN) [40], the generation process is conditioned on control input to the generator and discriminator (Figure 3 (b)). This provides additional information that helps the network to reproduce desired characteristics in the target class. From a simple, fully connected architecture in [33], [40], many new architectural innovations have been introduced to improve the GAN's ability to model data in image domains. Notable among these include the Deep Convolutional GAN (DCGAN), which employs convolutional layers for the generator and transpose-convolutional layers for the discriminator instead of fully connected layers throughout; the Laplace Pyramid GAN (LAPGAN) [41], which uses multiple generator-discriminator pairs in a multi-scale pyramidal structure; Information Maximizing Generative Adversarial Network (InfoGAN) [42], which proposes a non-conventional representation of the multi-dimensional noise, dividing it into an input noise vector and a latent variable, in order to help learn complex factors that control image appearance; Super Resolution GAN (SRGAN) [43], utilizes deep CNN together with adversarial networks to significantly increase the resolution training images; Pix2pix [40] propose paired image-based image-to-image translation approach using a cGAN model, DiscoGAN [44] and CycleGAN [45] both utilize a pair of generators and discriminators to perform unsupervised image-to-image translation (i.e., they convert images from one style to the another without using paired images); and Self-Attention GAN (SAGAN) [46], which incorporates attention mechanism into the GAN structure to model more global visual features and long-term dependencies. These and many other improvements have enabled GANs to synthesize high-detailed, photorealistic images with natural texture and lighting for training deep models. Figure 5 provides a visual illustration of how GANs have advanced over the years. Figure 6 shows how photorealistic images can be generated by modern generative adversarial networks (in this case BigGAN )can generate. In Figure 8, we show the multi-view 3D image synthesis capabilities of several state-of-the-art GAN models: GRAF [47], GIRAFFE [48], pi-GAN [49] and MVCGAN [50].

Some GAN models [51], [52], [53], [54] incorporate autoencoders into their structures to learn a latent space in which different images attributes (e.g., object texture, pose, or facial expression) can be easily manipulated. An autoencoder is basically made up of an encoder and decoder sub-models. The encoder transforms input data into a random, lower-dimensional representation in latent space. The decoder reconstructs the original data by mapping the low-dimensional data in the high-dimensional output space. The overall goal is to guarantee that the reconstructed data is as similar as possible to the original data. Models that employ autoencoders in their structure are able to learn a joint distribution of data and latent variables. Learning this joint space simplifies the process of creating desirable visual characteristics in output space as this can be accomplished indirectly by manipulating latent variables.

**b. Variational autoencoders**

Another class of generative AI techniques that has shown enormous promise is the variational autoencoder (VAE). Like the generative adversarial network, the variational autoencoder can be trained to generate novel data from a given domain. In addition to data generation, VAEs, like GANs, can perform tasks such as anomaly detection and correction, noise elimination, and data refinement. As we will see in Subsection 4.2, all these use-cases are important data augmentation functions in computer vision. Variational autoencoders are commonly used jointly with GANs and other generative models to achieve better results. They have been used to synthesize realistic images, videos as well as text for computer vision tasks.

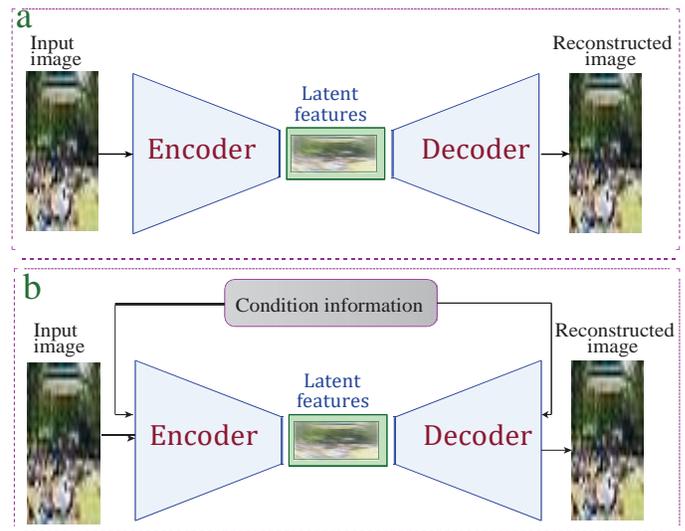

Figure 4. Functional block diagram of a basic VAE (a) and Conditional VAE (b) models.

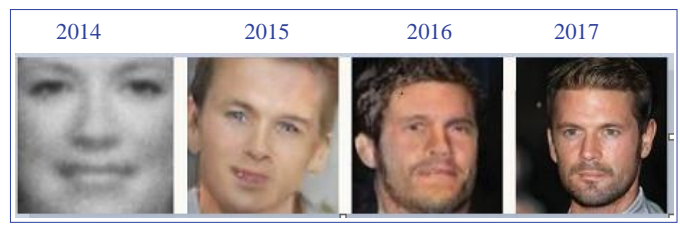

Figure 5. The rapidly improving performance of generative adversarial networks. Shown here is the quality of generated images by state-of-the-art models through the period 2014 to 2017. Images courtesy Brundage et al. [55].

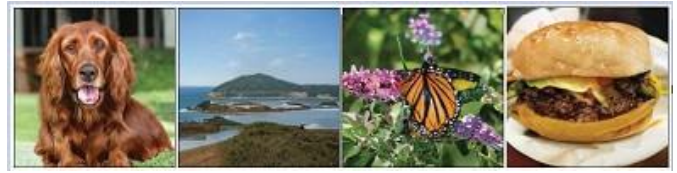

Figure 6. High quality photorealistic images generated by BigGAN [56].

The VAE (Figure 4a) was first proposed by Kingma and Welling in [34]. Since then, VAEs have been widely used to generate data for training deep learning models. Similar to GANs, VAEs learn the probability distribution of input data to help generate new data with similar characteristics. VAE first maps each input point to a normal distribution in a latent space using the encoder sub-model. The decoder then samples from this normal distribution to generate new samples, ensuring that the distribution of the real and generated data is as close as possible. The conditional variational autoencoder (cVAE), like the conditional GAN, uses additional metadata to generate outputs based on desirable characteristics defined by conditional input (Figure 4b). Since by their nature VAEs are not susceptible to the mode collapse problem that GANs suffer from, and GANs generate much perceptually-better images as compared to VAEs, many recent works (e.g., [37], [38], [39], [57] have proposed more complex generative modeling frameworks that leverage the advantages of both types of models.

## 4.2 Approaches to image data augmentation with generative AI models

In the context of data augmentation, generative models can be applied in several different ways. These include to generate new images, to transfer specific image characteristics from source to target images, and to enhance the perceptual quality or diversity of training data. data augmentation approaches based on these different principles have been used in diverse computer vision tasks, including medical image classification [58], [59], object detection [60], pose estimation [61] and visual tracking [62]. Common approaches to solving data augmentation problems based on generative modeling techniques are presented in subsections 4.2.1 to 4.2.4. The key aspects and application scenarios of these methods are summarized in Table 1.

### 4.2.1 Image synthesis

In application settings where it is difficult or impossible to obtain sufficient labelled data, the main goal of generative modeling is to generate synthetic data [63], [64], [65], [66], [67] to be used in place of, or in combination with real data. Models used in this situation are aimed at synthesizing specific categories of image data to aid training. The primary objective of the generative modeling, then, is to generate samples that cover the distribution of the underlying categories. This type of data can be achieved with conventional CNN-based GAN architectures without utilizing conditional information as in the case of conditional GANs or conditional VAEs. For example, Kaplan et al. [68] demonstrated the ability of GAN and VAE to generate photorealistic retinal images without using conditional information. In [64], Bowles et al. employ a PGGAN model to synthesize images for Computed Tomography (CT) and Magnetic Resonance (MR) image segmentation tasks. The authors showed that using GANs to generate synthetic images improves segmentation performance on the two different tasks, irrespective of the original data size and the proportion of synthetic samples added. To improve the perceptual quality of generated images, several GANs may be used, each tuned for creating specific category (e.g., in [63]). In [66], Souly et al. proposed generative adversarial network to generate large amount of labeled images in a semi-supervised manner using unlabeled GAN-synthesized image data to help in semantic segmentation tasks.

### 4.2.2 Image-to-image translation

Image-to-image translation [40] is a technique used to transform an image by transferring its content into the visual style of another image. In its basic form, the approach involves learning a mapping from source to target domain.The approaches rely on principles of conditional generation using models such as cGANs and cVAEs. Generative modeling methods based on image-to-image translation can be used to convert images from one color space to another. In particular, approaches for converting among infrared, grayscale and RGB color images are common [73], [74]. The techniques can also enable different visual effects and specific features such as contrast, texture, illumination and other complex photometric transformations which would otherwise be challenging for traditional augmentation approaches. The approach, as a data augmentation method, has a wide scope of applications in computer vision. In medical imaging applications, for example, approaches based on image-to-image translation can be used to transfer images from one modality to another (e.g., from CT to MRI or X-ray image format). Figure 7 shows different translations using StyleGAN model [45].

Another important application of image-to-image translation is to synthesize view-consistent scenes, where 3D information and the overall spatial structure of the scene is preserved in the process of translation [49], [75]. This is extremely useful in semantic scene understanding tasks. Image-to-image translation approaches have also been widely used to generate images of novel views from particular views such as frontal facial views from angular views (e.g., [76], [77]) or to produce different human poses from a single pose (e.g., [78]). The effectiveness of image-to-image translation as a viable data augmentation strategy has also been demonstrated in challenging computer vision tasks such as visual tracking [79], [80], [81], person re-identification [82], [83], [84], object detection under severe occlusion [85] and strong lighting conditions [86]. Figure 8 depicts multi-view images generated by different GAN models.

While traditional image-to-image translation approaches [40] are generally based on cGANs and cVAEs architectures that utilize paired images, new techniques are based on the concept of cyclic consistency. For example, CycleGAN [45] DualGAN [87] and DiscoGAN [44] can translate images from one style to another without paired images. These methods learn a mapping function between source and target image domains by means of unsupervised learning — that is, images in the target domain do not have corresponding examples in the source. The approach is useful in many practical application since it is often challenging to obtain paired images in real-world scenarios for training machine learning models. For instance, for a specific environment, images for autonomous driving tasks may only be available for a limited set of weather conditions, but it may be required to improve robustness by training on a wide range of possible conditions. In such a situation, with unpaired image-to-image translation methods, the available content image can be transferred to all desired visual appearances without requiring additional content images.

The integration of large language models (LLMs) [88] and vision-language model (VLMs) [89] with GANs, VAEs and other generative models allows the process of image-to-image translation to be automated using textual prompts. LLMs and VLMs can also enable textual descriptions of visual scenes to be automatically generated as supplementary input for training computer vision models.

### 4.2.3 Image manipulation

Another common generative modeling approach to data augmentation is to qualitatively transform training data in desired ways by performing specific photometric (e.g., [91], [92]) and geometric ( [71], [93], [94], [95]) image manipulations. Photometric image operations such as binarization

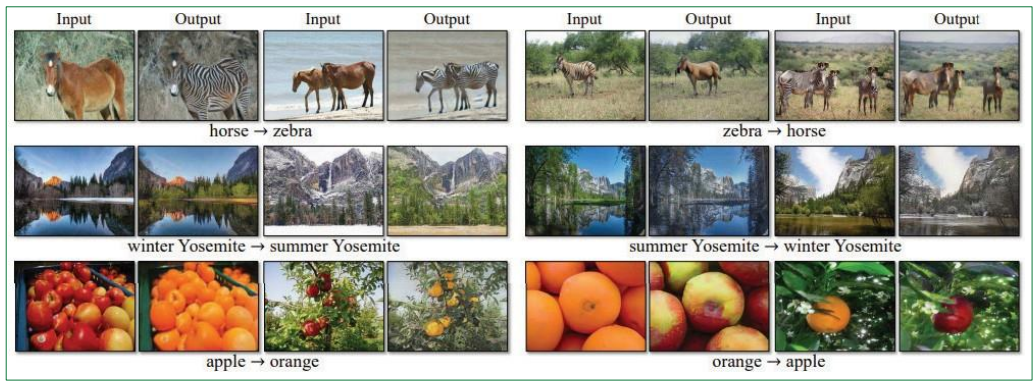

Figure 7. Examples of image-to-image translation by CycleGAN [45].

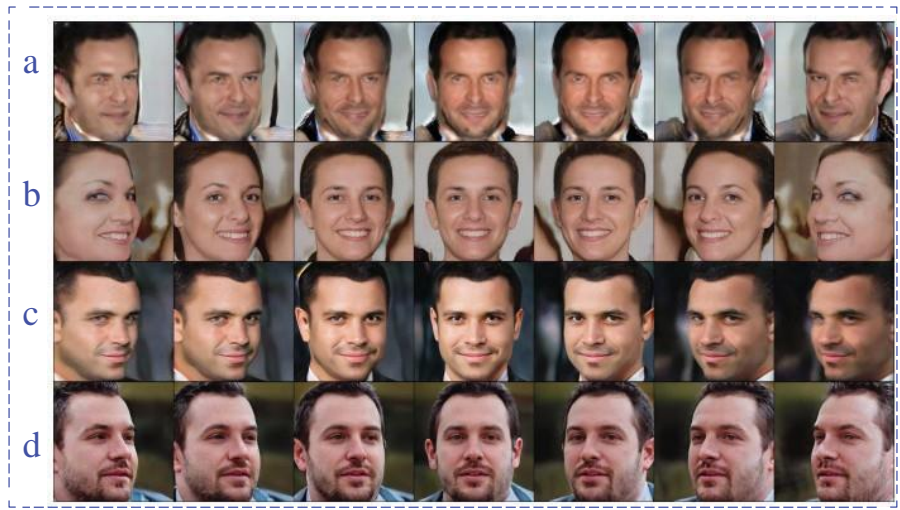

Figure 8. Photorealistic, multi-view images generated by 3D-aware image synthesis approaches: (a) GRAF [47], (b) GIRAFFE [48], (c) pi-GAN [49] and (d) MVCGAN [50].

Table 1
Common approaches to data augmentation by using generative modeling

| Approach | Main function of generative model | Classic models | Application scenario |
|---|---|---|---|
| Image synthesis | Generate new samples per given categories | Deep Convolutional GAN (DCGAN) (e.g., in [69]) | Synthesize new samples where no training data exist |
| Image-to-image translation | Reproduce specific visual characteristics | Conditional GAN (CGAN) and variants such as DiscoGAN [44] and CycleGAN [45] | Transfer desired visual appearance to training samples |
| Image manipulation | Perform specific transformations on input images | Spatial Transformer GAN (ST-GAN) [70]; MOST-GAN [71] | Produce diversity in training data to enhance generalization |
| Image Enhancement | Denoise images or improve quality or photorealism of training data | Denoising GAN (DN-GAN) [72]; Super Resolution GAN (SRGAN) [43] | Improve the perceptual quality of training samples |

[91], colorization (i.e., conversion from grey-scale to color images) [96] and dehazing [92] are common tasks that can be accomplished with generative modeling.

### 42.4 Image quality enhancement

In some computer vision tasks, images available for training deep learning models are often of low quality. One way to improve performance is to enhance the quality of the training data. For example, generative modeling is commonly used to clean noisy images (e.g., in [72]). Also, low resolution images, can be qualitatively improved by using super-resolution GANs such as Pix2Pix [40], SRGAN [43], ESRGAN [97] or their derivatives. In a recent work [98], Wang et al. used a modified Pix2Pix model as a super-resolution GAN to increase the resolution of low-resolution, microscopic images for training deep neural networks. They first generated additional data using CycleGAN before employing the super-resolution GAN to improve the quality of the training dataset. In many studies, generative models have been used to generate large, clean images from noisy (e.g., [99], [100]) data, low resolution images (e.g., [101]), corrupted labels (e.g., [102] ) or images taken in adverse weather conditions such as rainy weather [90]. Figure 9 shows the effectiveness of image enhancement techniques such as de-raining in improving the perceptual input data. Generative modeling approaches that apply geometric transforms on training samples have also been reported in [103], [104], [105]. Some recent approaches are aimed at enhancing the perceptual quality of CAD-generated models. For example, RenderGAN [105] and DA-GAN [106] seek to improve performance by refining simple, synthetically generated 3D models so as to endow them with photorealistic appearance and desirable visual properties.

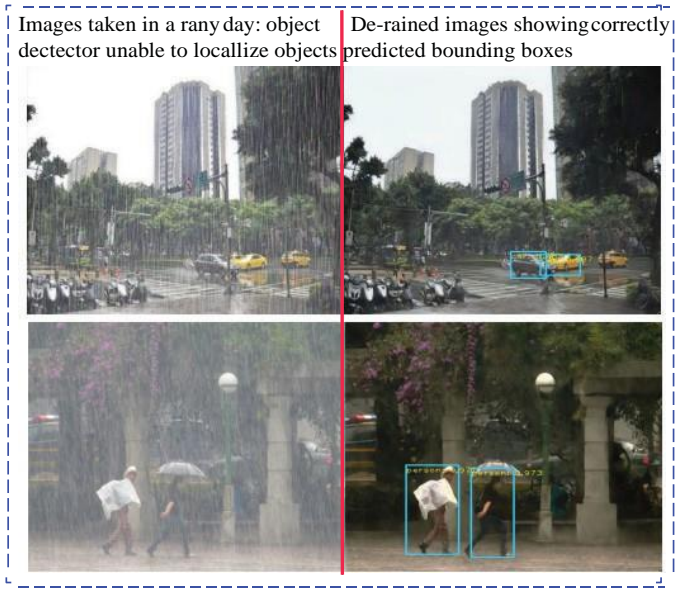

Images taken in a rany day: object detector unable to locallize objects | De-rained images showing correctly predicted bounding boxes

Figure 9. Examples of using generative modeling approaches to augment data by enhancing perceptual quality, in this case – by de-raining. The images by Zhang et al. [90] shows improvement in object detection performance after de-raining input samples.

### 4.3 Common limitations of generative modeling techniques and possible workarounds

One of the main problems with generative modeling methods is that they require very large training data for good performance [107]. GANs are also susceptible to overfitting – a situation where the discriminator memorizes all the training inputs and no longer offer useful feedback for the generator to improve performance. To address these problems of GAN performance, a number of works [108], [109], [110] have considered augmenting the data on which the generative model is trained. These approaches have been shown to be effective in alleviating small data and overfitting problems. However, employing augmentation strategies can lead to a situation where the generator reproduces samples from the distribution of the augmented data which may not be truly representative of the target task. Consistency regularization [111], [112] is a recent approach that has been proposed to prevent augmented data from being strictly reproduced by the generator. More advanced methods to improve GAN generalization include techniques based on perturbed convolutions [113] and Extreme Value Theory [114]. For instance, to enable GANS to handle rare samples, Bhatia et al. [115] proposed a probabilistic approach based on Extreme Value Theory [114] that allows to generate realistic as well as out-of-distribution or extreme training samples from a given distribution. In this context, extreme samples are training examples that deviate significantly from those present in the dataset. The approach also provides a way to set degree of deviation and the likelihood of the occurrence or proportion of these deviations in the generated data. Liu et al [116] demonstrated that GAN may in some situations fail to generate the task-required data (as a result of optimizing for a different task altogether) because GANs may be optimizing for a different objective. Specifically, in [116] a GAN designed for object detection tasks was shown to (have) rather optimize for realism of generated images.

Like with all data synthesis methods, it is currently not possible to directly compare different sets of synthetic data, or even to determine the suitability of synthesized data for a particular task without carrying out exhaustive tests. The fundamental problem is the general lack of quality metrics that can objectively evaluate the fitness of data for a given task. While techniques like log-likelihood provides a means to evaluate and assess the quality of VAEs, it is currently difficult to extend these to objectively compare the quality of GAN models. Some workarounds exist that allow to roughly estimate the quality of generated samples based on their similarity with the target data. This involves comparing the statistics of the generated data to that of the target data. The simplest metrics involve using more traditional similarity measures such as nearest neighbors, log-likelihoods [57], Minimum Mean Discrepancy (MMD) [117] and Multi-scale Structural Similarity Index Measure (MS-SSIM) [118]. Since these techniques merely estimate pixel distribution, high scores on the metrics do not strictly indicate high image quality. More advanced metrics allow to quantitatively estimate data diversity (i.e., the degree to which the synthetic data approximates the distribution of the target data), quality (i.e., overall photorealism), and other characteristics. The most important of these metrics include Inception Score (IS) [119] and Fréchet Inception Distance (FID) [120], as well as their new variants such as Spatial FID [121], Unbiased FID [122], Memorization-informed FID [123] and Class-aware FID [124]. These metrics allow to evaluate not only the general quality but also important aspects such as bias and fairness of generative models. Manual evaluation by visual inspection is another common way to determine the quality of data [125], [126] synthesized using generative modeling techniques. The approach relies on the developer's domain knowledge to make good judgment about the appropriateness of the training data. In some cases, this may offer the best guarantee for success. However, the approach is very subjective and prone to biases of the human assessor. Moreover, because of the limited capacity of human experts, the method cannot be applied in settings that involve large-scale dataset.

A common class of problems with generative modeling techniques relates to training challenges. In particular, generative models based on GANs suffer from unstable training. One of the main causes of this issue is the so-called mode collapse problem [127]. This phenomenon occurs when the generator fails to learn the variety in input data and is, thus, able to generate only a particular type of data that consistently beats the discriminator but is inferior in terms of diversity. Common solutions to this problem include weight Normalization [128] and other regularization techniques [129] as well as architecture innovation [130]. Another serious problem with training generative models is the non-convergence problem [131]. Researchers have attempted to address this difficulty by employing techniques such as adaptive learning rates [132], [133], restart learning [134] and evolutionary optimization of model parameters [135]. These approaches alleviate the problem to an extent but do not completely eliminate it.

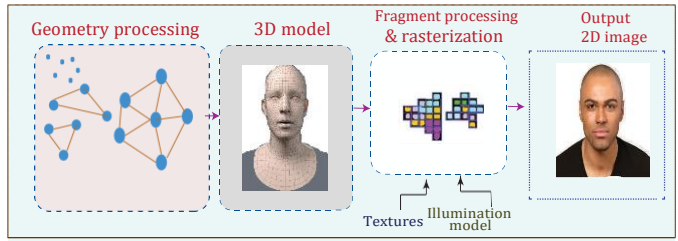

Figure 10. Simplified pipeline of the graphics modeling process.

## 5 Computer graphics modeling

An increasingly promising line of works [136], [137], [138] that aims to address the data scarcity problem exploit computer graphics tools to synthesize training data. Computer graphics tools are capable of creating 2D and 3D objects as well as whole complex scenes. The procedure for synthesizing data using computer-aided design (CAD) techniques involves complex processes such as modelling, rigging, texturing, and animating of the generated 3D objects. Game engines provide more advance modeling capabilities that can be used to create large, interactive scenes and virtual environments that span whole cities.

### 5.1 Approaches to data synthesis based on computer graphics modeling

In this subsection, we discuss the various graphics modeling methods commonly used to generate synthetic to address augmentation problems.

#### 5.1.1 CAD modeling

The methods discussed in Section 4 typically tackle the data augmentation problem as a 2D mapping of a particular image domain to itself by using various transformations to re- create variations of the original data in 2D space. These 2D-based transformation approaches lack semantic 3D grounding, and are, thus, highly superficial in nature and may not adequately represent the actual variations of real-world scenes. Graphics modeling approaches address this limitation by approaching data augmentation as a 3D to 2D mapping (i.e., a mapping from 3D physical world to 2D pixel representation). One important area where computer graphics models are increasingly used is in the area of 3D perception. By modeling the underlying 3D processes of image formation, simulation-based methods produce qualitatively better augmentations compared to pure 2D manipulation methods (for 3D vision). In particular, performance on tasks such as pose understanding, gesture and action recognition are immensely aided by 3D supervision. Techniques based on CAD modeling can also simulate nonstandard visual data such as point clouds (e.g., [139]), voxels (e.g., [140]), thermal images (e.g., [141]), or a combination of two or more of these modalities (e.g., clouds [142]). State-of- the-art computer graphics tools are able to produce fairly realistic visual data for training machine learning models. Three-dimensional game engines are particularly promising in this regard, as they can simulate complex natural processes and generate near-realistic environments under different conditions using real physics models. This capability provides an opportunity to train machine learning models on complex real-world (natural) scenes. Examples of simple 3D objects from the Amazon Berkeley Objects (ABO) dataset modeled using CAD tools are shown in Figure 11. Figure 12 shows realistic indoor scenes from the Hypersim [143] dataset generated, also by CAD tools.

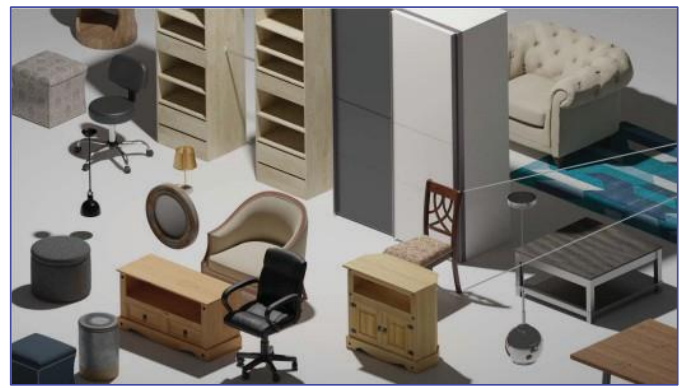

Figure 11. Sample objects from the Amazon Berkeley Objects (ABO) dataset [144]. It is an example of dataset constructed using 3D CAD models. The dataset consists of a large collection of artistically created 3D models of common household objects, and includes the necessary metadata and physically-grounded properties.

CAD modeling methods represent basic scene information using geometric primitives such as triangles and polygons along with their location information and camera properties, global scene information. A renderer translates this representation into a complete 3D scene. Rendering is the process by which images are generated using these basic geometric primitives and additional scene parameters and textures. A typical rendering pipeline is a multi-stage process that composes the primitive geometric entities and scene parameters into more complex objects and scenes (Figure 10). The process involves sequentially creating higher-level representations at every processing stage assembling basic entities into more complex objects until whole scenes are created. The basic elements are specified using analytical formulas. A rendering engine translates the mathematical formulas into their corresponding graphical representations and compute various characteristics such as lighting, colors, and shadows for display. In an OpenGL rendering pipeline, for example, the rendering task consists of vertex shading and assembly — processes that define coordinates posisitions of objects and their attributes and their composition into higher geometric shapes like polygons; geometry shading and rasterization — i.e., converting the geometric informa-tion into pixel form; fragment shading – for processing color and texture information. Many advanced integrated development environments (IDEs) have been developed to facilitate 3D modeling. They provide advanced, intuitive graphics user interface (GUI) and easy-to-use toolsets for rendering and editing 3D models. Common 3D modeling tools include simulation and 3D animation software tools such as Cinema 4d, Blender, Maya and 3DMax. These tools provide a means to obtain task-specific data in situations where available data does not meed the requirements of the target task. For instance, Hattori et al. [145] employ 3DMax to synthesize data for human detection tasks in video surveillance applications where task-specific data may not readily be available. The approach allows the generated data to be customized according to the specific requirements of a scene (e.g., scene geometry and object behavior) and surveillance system (i.e., camera parameters). The tools have also provided a means to create large-scale datasets for generic applications. Examples of large-scale synthetic datasets obtained from 3D CAD models include ShapeNet [146], ModelNet [147] and SOMASet [148] datasets. Some of the most important datasets created using 3D modeling tools are described in Section 8.

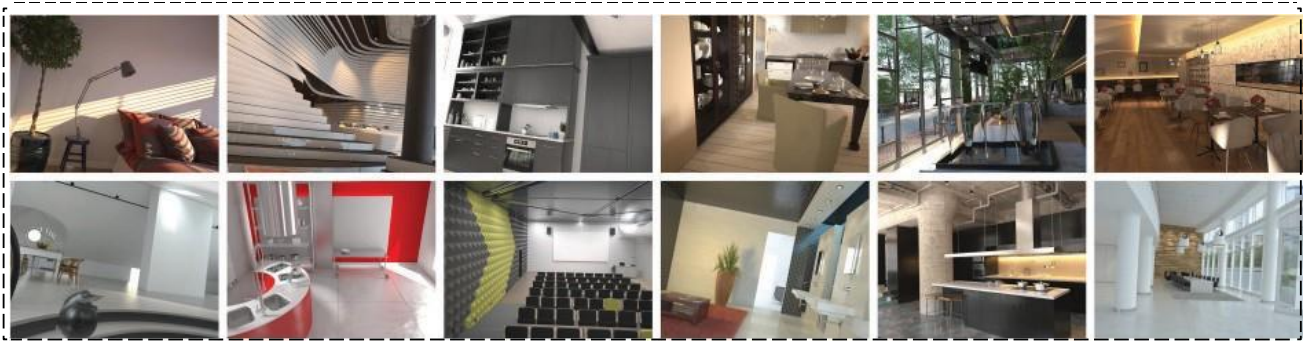

Figure 12. Examples of photorealistic synthetic indoor scenes from the Hypersim [143] dataset.

### 5.1.2 Synthetic data from 3D physics (game) engines

While CAD tools are primarily used for creating 3D assets, game engines provide tools to manipulate the generated 3D objects and scenes in nuanced ways within virtual environments. They typically come with built-in rendering engines like Corona renderer, V-ray and mental ray. Advanced game engines such as Unity3D, Unreal Engine, and Cry Engine can simulate real-world phenomena such as realistic weather conditions; fluid and particle behavior, effects including diffuse lighting, shadows and reflections; object appearance variations resulting from the prevailing phenomena. By randomizing parameters associated with these phenomena, sufficient data diversity can be achieved. Besides visual perception, simulated environments based on 3D game engines can serve for a broad range of applications. They are particularly suitable for training models in domains like planning, autonomous navigation, simultaneous localization and mapping (SLAM), and control tasks. Figures 13 and 15 show sample scenes from Carla [1] and AirSim [149], [150], respectively. Both tools are created from Unreal Engine. Figure 14 shows the different sensing modalities that can be obtained from Carla.

Because of the advanced manipulation capabilities of modern game engines, recent synthetic data generation approaches [151] favor the use of 3D game engines, which are capable of generating complete virtual worlds for not only training neural network models, but also enabling interactive training of elements of worlds using deep reinforcement learning frameworks. For instance, ML-agents introduced in the Unity 3D game engine, provides a framework for training intelligent agents in both 2D and 3D worlds using a variety of machine learning techniques, including imitation learning, evolutionary algorithms and reinforcement learning.

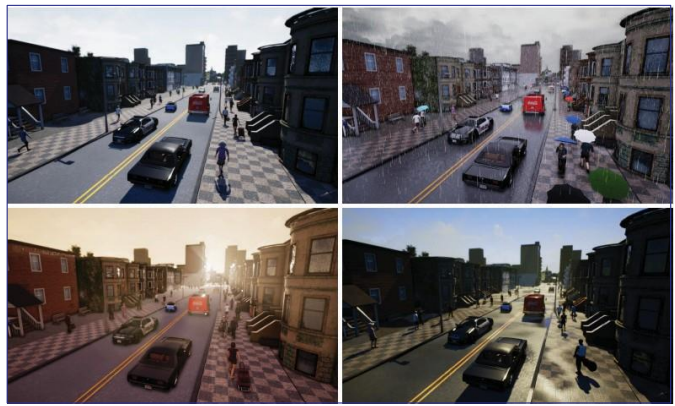

Figure 13. Varying appearance of a sample synthetic scene from CARLA simulator [1] in different weather conditions.

Varol et al. [3] synthesized realistic datasets using Unreal Engine. Their synthetic dataset, synthetic humans for real tasks (SURREAL) has been provided as open-source dataset for training deep learning models on different computer vision tasks. The authors in [3] showed that for depth estimation and semantic segmentation tasks, deep learning models trained using synthetic data generated by 3D game engines can generalize well to real datasets. Jaipuria et al. [152] used Unreal Engine to enhance the appearance of artificial data for lane detection and monocular depth estimation in autonomous vehicle navigation scenarios. As well as generating scenes with photorealistic, real-world objects, they also simulated diverse variability in the generated data: viewpoints, cloudiness, shadow effects, ground marker defects and other irregularities. This diversity has been shown to improve performance under a wider range of real-world conditions. Bongini et al. [141] rendered synthetic thermal objects using U3D's thermal shader and superimposed them in a scene captured using real thermal image sensors. They additionally employ a GAN model to refine the visual

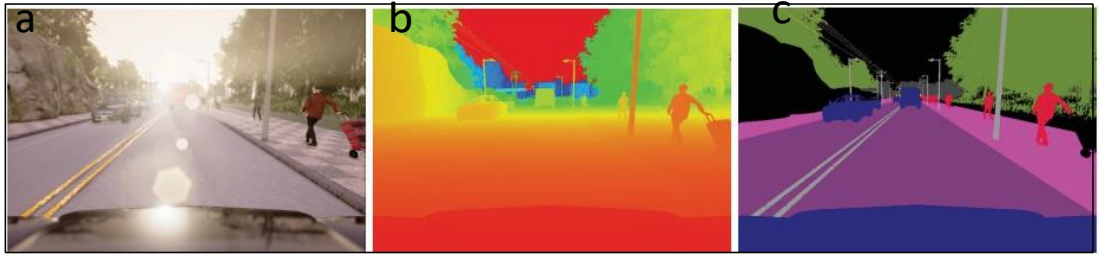

Figure 14. CARLA provides three sensing modalities: traditional vision (a), depth (b), and semantic segmentation (c).

appearance of the rendered images so that they look more like natural thermal images.

Additional plugins have been developed to facilitate the ease of generation of image data and corresponding labels from game engines [153] or from virtual environments [153], [154] developed on the basis of 3D engines. For instance, Borkman et al. [153] introduced a Unity engine extension known as Unity Perception that can be used to generate artificial data and the corresponding annotations for different computer vision tasks, including pose estimation, semantic segmentation, object detection and image classification. The extension has been designed to synthesize data for both 2D and 3D tasks. Hart et al. [155] implemented a custom OpenCV tool in Robot Operating System (ROS) environment to extract frames from simulated scenes in Gazebo platform and generate their corresponding labels. Similarly, Jang et al. [154] introduced CarFree, an open-source tool to automate the process of generating synthetic data from Carla. The utility is able to generate both 2D and 3D bounding boxes for object detection tasks. It is also capable of pixel-level annotations suitable for scene segmentation applications. Carla [1] provides a python-based (API) for researchers and developers to interact with and control scene elements. Mueller and Jutzi [156] utilized Gazebo simulator [157] to synthesize training images for pose regression task. Kerim et al. [158] introduced the Silver framework, a Unity game engine extension that provides highly flexible approach to generating complex virtual environments. It utilizes the built-in High Definition Render Pipeline (HDRP) to enable control of camera parameters, randomization of scene elements, as well as control of weather and time effects.

### 5.1.3   Obtaining data directly from video games

Some recent works, for instance [159], [160], [161]), have focused on directly extracting synthetic video frames from scenes of video commercial games as image data for use in training computer vision models.. To achieve this, appropriate algorithms are used to extract and label random frames of video sequences by sampling RGB images at a given frequency during game play. Since modern video games are already photorealistic, the qualitative characteristics of image data obtained this way is adequate for many computer vision tasks. Shafaei et al. [159] showed that, for semantic segmentation tasks, models trained on synthetic image data obtained directly through game play can achieve comparable generalization accuracies as those trained on real images. Further refinements by means of domain adaptation techniques to bridge the inherent seman-tic gap between real and synthetic images results in better performance than models trained on real image datasets.

Since it is generally more challenging to obtain relevant data for video object detection and tracking tasks than for other computer vision applications, generating data from video games has emerge as a promising workaround to alleviate the challenge (e.g., in [161], [162], [163], [164]). The main advantage of this approach is the possibility of utilizing off-the-shelf video games without strict requirements for the resolution of the captured images. Its main disadvantage is that video games contain general environments that are not tailored for specific computer vision applications. Also, the visual characteristics of images from game scenes may not be optimized for computer vision tasks. Moreover, there is generally a lack of flexibility in the data generation process as the user can exercise very little control over the scene appearance and cannot change scene behavior as needed; since scenes in the video sequences are fixed, users are not able to introduce factors of variability (e.g., arbitrary backgrounds, objects and appearance effects) into the scene. In contrast, approaches such as [158], [163], [165], [166] that synthesize scenes from scratch using 3D physics engines bypass these limitations, but require enormously long time and are labor-intensive.

### 5.1.4   Obtaining synthetic 3D data through scanning of real objects

Another technique [138] to alleviate the laborious work required in synthesizing dense 3D data from scratch using graphics modeling approaches is to leverage special tools such as Microsoft Kinect to capture relevant details of the target objects. This is accomplished by scanning different views of the relevant objects at different resolutions and constructing mesh models from these scans. With this approach, the desired low-level geometric representation can be obtained without explicitly modeling the target objects. In the simplest case, the 3D representation can be obtained with depth cameras to capture multiple views of the object. Suitable algorithms such as singular value decomposition (SVD) [168], random sample consensus (RANSAC) [169] and particle filtering [168] are then used to combine these multiple images into a composite 3D model. The approach proposed in [169], [170], [171] utilize RGBD cameras as 3D scanners for extracting relevant appearance information from object. Figure 16 shows sample frames from the Open-Rooms dataset [167], a dataset created from 3D-scanned indoor scenes—ScanNet [172].

A common practice is to construct basic articulated 3D models from the 3D scans which are further manipulated within 3D modeling environments to create variability. The approach allows to incorporate only the relevant objects into already modeled 3D scenes. The use of real 3D scans of objects also helps to achieve physically-plausible representations of the relevant objects. Furthermore, visual effects such as lighting, reflections and shadows cannot easily be manipulated by conventional 2D image transformation methods. Therefore, these techniques are vital in situations where it is necessary to train deep learning models to be invariant with respect to these visual phenomena. For instance, Chogovadze et al. [173], specifically employ Blender-based light probes to generate different illumination patterns to train deep learning models robust to illumination variations. Vyas et al. [169] employed a RGBD sensor to obtain 3D point clouds and then used a RANSAC-based 3D registration algorithm to construct the geometric representation from the point cloud data. The authors obtained an accuracy of 91.2% on pose estimation tasks when trained using the synthetic dataset, albeit with some domain adaptation applied. It must, however, be noted that this method can only be used in situations where access to the target domain data is possible.

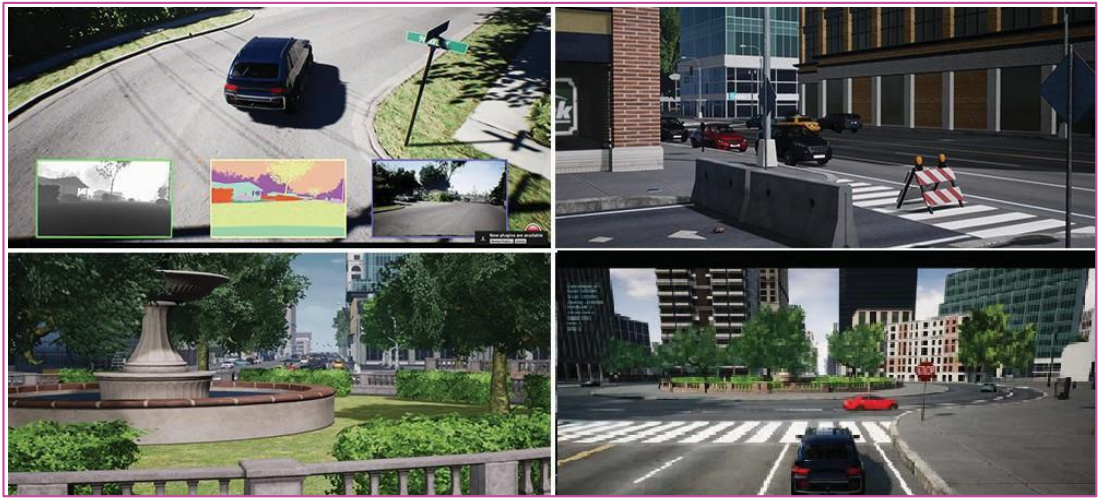

Figure 15. Sample scenes from Microsoft AirSim [149], a synthetic virtual world for training unmanned aerial vehicles (UAVs) and autonomous ground vehicles. The views show typical urban environments for autonomous driving.

### 5.1.5  Combining synthetic with real data

Because of the complex interaction of many physical variables which are difficult to capture using computer graphics methods, some researchers suggest using synthetic data simultaneously with real data. A few works suggest using synthetic data only as a means for defining useful visual attributes to guide the augmentation process. In [174], for instance, Sevastopoulos et al. propose an approach where synthetic data from a Unity-based simulated environment is used as the first stage of data acquisition process to identify useful visual attributes that can be exploited to maximize performance in a given task before collecting real data. The idea is to leverage synthetic data to provide initial direction for further exploration so as to lower the cost of excessive trial and error experimentation on real data. In Section 6, we present quantitative results on the effectiveness of data augmentation approaches that combine real and synthetic data.

### 5.1.6  Integrating real and virtual worlds

As discussed earlier, the basic idea of the 3D object scanning approach is to obtain information about target objects as stand-alone data by extracting 3D information of real objects and then utilizing graphics processing pipelines to perform 3D transformations on the skeletal models obtained by scanning. However, in some cases (e.g., [175], [176]), 3D geometry models of objects obtained by scanning the real-world objects are incorporated into more complex, task-relevant 3D scenes created using graphics tools. Integrating scanned objects into such virtual worlds provide effective means to manipulate and randomize different factors so as to simulate complex real-world behavior useful for model generalization and robustness. Synthetic 3D models obtained in this manner can be used to train models on complex 3D visual recognition tasks such as object manipulation and grapping. They can also be rendered as 2D pixel images to augment image data. In some other cases (e.g., [177]), synthetic objects are immersed into real worlds in augmented reality fashion. These real worlds are obtained from camera images or videos. In [177], this approach was shown to outperform techniques based on purely real or synthetic data. The most important data augmentation approaches based of computer graphics modeling are summarized in Table 2.

## 5.2  Challenges of computer graphics-based synthesis and workarounds

Despite the advanced rendering capabilities of modern 3D modeling tools, the use of graphics modeling tools to synthesize training data has a number of limitations:

- The synthesis of realistic data with high level of detail and natural visual properties including realistic lighting, color and textures is a complex and time-consuming process. This may limit the scope of applications of 3D graphics modeling techniques in data augmentation applications to only simple or moderately complex settings.
- Currently, in many situations, there is no objective means of assessing the quality of artificially gener-

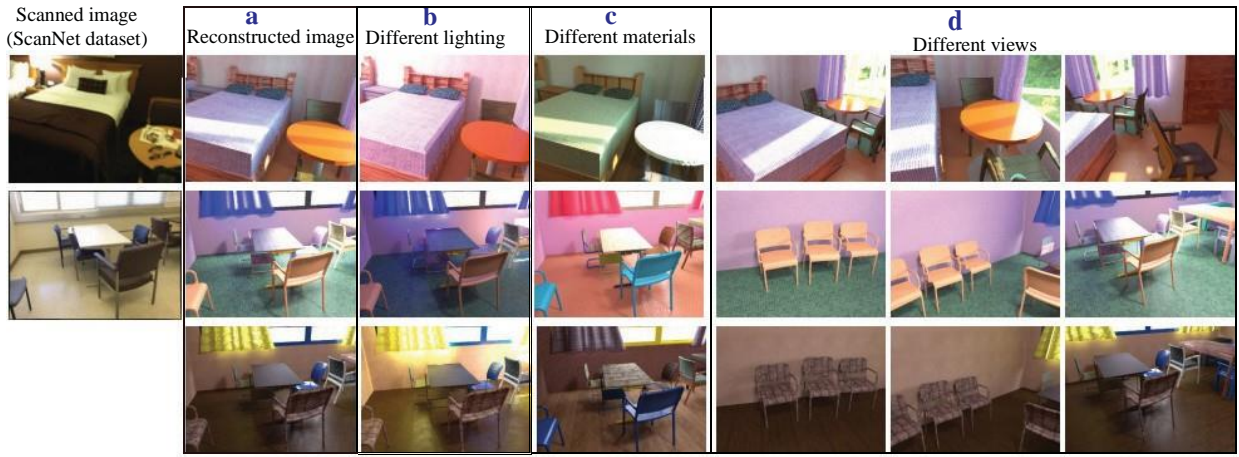

Figure 16. Scanned indoor scenes from ScanNet are used to generate new synthetic 3D data based on the approach in [167]. The reconstructed synthetic scenes (a) can then be rendered with different illumination levels and patterns (b), different materials (c), or different views (d).

Table 2
Summary of data augmentation approaches based on computer graphics modeling techniques

| Approach | Description | Sample works |
|---|---|---|
| CAD modeling | Simulates 3D objects using geometric primitives | [146] |
| 3D physics engines | Allows to generate and manipulate physically-plausible 3D objects in virtual environments | [3] |
| Data from gameplay | Generates pseudo-video data directly from video games | [163] |
| Data from scanners | Produces synthetic 3D data through scanning of real objects | [138] |
| Real-virtual world integration | integrates scanned 3D objects with CAD or physics engines | [175] |

ated data. While some methods have been devised to evaluate synthetic data, these are only applicable in situations where reference images are available for comparison. However, in many of the settings that necessitate the use of graphics models, samples of the target images are generally not available at the model development and testing stages.

- It is often difficult to know beforehand the desirable factors and visual features that are good for performance. Moreover, synthetic data that appear photorealistic to the human observer may not actually suitable for a deep learning model. For instance, while high-fidelity synthetic samples have failed to provide satisfactory performance in some situations, some researchers (e.g., in [160], [178], [179] have obtained good performance using low-fidelity synthetic images.

- The generation process is usually accomplished by a careful modeling process, and not from any natural processes or sensor data obtained from real-world variables. Even with the methods that generate synthetic data by scanning real objects, 3D alignment techniques used for registration also introduce additional imperfections into the scene representation. All these problems exacerbate domain gap problem between real and synthetic data.

As a result of the above limitations, it is often challenging to produce semantically meaningful synthetic environments comparable to natural settings, especially when dealing with complex scenes. Since simulated data obtained by 3D modeling tools are often not perfect, the process of augmentation does not only consist in modeling visual scenes, but also in correcting various imperfections and refining the appearance of the artificially generated data to mitigate the domain gap between synthetic and real data. Indeed, a large number of so-called sim-to-real (sim2real) techniques [180], [181], [182], [183] have been proposed for fine-tuning and transferring synthetically generated graphics-based data to real-world domains. The use of generative modeling techniques (e.g, [105], [184], [185]) as a means to endow simple 3D models with photorealistic appearance and desirable visual characteristics has recently gained attention. These approaches provide a more practical and cheaper means for generating extra training data by leveraging unlabeled image data and GANs to introduce hard-to-model, real-world visual effects to simple computer graphics images. In [105] Sixt et al. proposed to learn augmentation parameters to enhance the photorealism of synthetic 3D image data using a large set of unlabeled, real-world image samples. Dual-agent generative adversarial network (DAGAN) has been proposed to enhance the photorealism of synthetic, rudimentary facial data generated by 3D models [106]. Atapour and Breckon in [186] employ a CycleGAN model to refine the appearance of synthetic data which resulted in improved performance. Instead of employing GAN to perform visual style transfer in order to refine synthetically generated data, Huang and Ramanan in [187] propose to produce a rather large volume of synthetic pedestrian data and then select the most realistic poses through adversarial training.

Rich geometric features inherent in synthetic data have also been used to refine simple 2D images. An important class of refinement methods relies on image-to-image translation models to transfer novel appearance and geometry information (e.g., new poses of objects, occlusion patterns, etc.) from data sources where these views are prevalent to new datasets. For instance, Liu et al. [82] utilize a GAN model to increase the diversity of human poses in a new dataset by transferring pose information from the motion analysis and re-identification set (MARS) dataset [188] where rich pose variation is present. In contrast, some approaches [189], [190] propose to explicitly perform the necessary transformation analytically. Ma et al . [190] employ a deep neural network architecture that combines a GAN and a U-Net sub-models to generate new training samples conditioned on person images and the corresponding poses. The U-Net model reproduces new images of the reference image in the desired poses by explicitly utilizing the specified pose information. A GAN model is then used to correct any artifacts that may result and to refine the general appearance of the person in the desired pose. Similarly, Chen et al. [189] employ explicit warping operations to synthesize new shapes and poses for pedestrian datasets. The concept of transferring realistic views to more rudimentary visual data is extremely useful in person or pedestrian detection [187], [191] and tracking [80], [81], as well as person re-identification tasks [82], [83], [84].

### 5.3 Augmentation for tasks not requiring realistic data

As noted in the preceding paragraphs, many recent image synthesis approaches focus on generating realistic data for training machine learning models. This process is often tedius and sometimes requires expensive third-party tolls to acomplish. However, in some cases –for example, depth perception [192], optical flow [179] and ego-motion estimation [193], disparity learning for binocular (stereo) vision [194], [195] – the perceptual quality of images is not important for visual understanding. Since the problem of modeling 3D data with realistic geometry and appearance is often challenging, models designed to solve these tasks can easily leverage easy-to-model, non-photorealistic 3D data synthesized with computer graphics tools. Shotton in [192] trained a machine learning model on non-photorealistic 3D data to generate depth maps for human posture recognition. The model produced good results when trained on perceptually poor quality synthetic data alone. Fischer [179] trained a deep learning model (FlowNet) on the Flying Chairs dataset, a collection of unrealistic 3D models of chairs, for optical flow. Ilg et al., observed in [196] that the less realistic flying chairs dataset produces better results in optical flow estimation than [194], which contains more photorealistic 3D training data. The not-so-realistic FlyingThings3D dataset [194] has also proven effective in training deep learning models for optical flow and scene flow tasks [193], [195].

## 6 NEURAL RENDERING

Another common way to synthesize new training data for visual recognition tasks is by neural rendering. The aim of neural rendering is to realize the scene rendering process using deep learning models. Unlike traditional scene rendering based on 3D graphical modeling, the neural rendering process can be accomplished in both forward and backward directions. In the forward direction, 2D images are generated from 3D scenes and additional scene parameters. In the backward direction, the pixel image is translated into a realistic 3D scene. The pipeline for this process is depicted in Figure 17.

### 6.1 Differential neural rendering

Because the rendering process is inherently non-differentiable, its incorporation in deep neural networks is severely constrained. Differentiable rendering is an approach to overcome this challenge by formulating the scene modeling process as a differential problem that can be seamlessly incorporated into deep neural network pipelines and trained end-to-end. To achieve this, numerical techniques have been proposed to find approximate derivatives of the rendering operations that can be used in gradient-based algorithms such as backpropagation to optimize scene parameters. Some earlier works such as [198] and [199] embed real 3D graphics components as renderers for the forward rendering, and accomplish reverse rendering by modeling the relationship between the output 2D image and the input scene parameters using approximate analytical functions. An important task in rendering is rasterization – i.e., converting scene representation from continuous raster form into discrete pixel values. Since the rasterization process is inherently non-differentiable, many of the current approaches are focused on developing approximating methods that can handle this process in a differentiable way. Indeed, a large number of differential neural rendering approaches usually restrict the differentiation to rasterization process only. Typically, the process involves estimating the gradients of the rasterization process and using it in a back propagation algorithm to optimize parameters for the rendering task.

Recent works [200], [201], [202] have focused on learning many of the stages of the forward rendering process in an end-to-end manner using deep learning techniques. The use of deep neural network models in the rendering process can help to generate more nuanced scene attributes (objects, environments, realistic textures and noise) or refine existing scene elements to improve the performance of computer vision models when trained on synthetic data.

### 6.2 Scene representation methods

An important part of the neural rendering process is the representation of scene elements: geometric priors and scene parameters. We discuss the common approaches for scene representation. The main strengths and weaknesses of these approaches are summarized in Figure 18.

#### 62.1 Explicit representation of scene primitives

Traditional computer graphics methods use explicit representations that leverage analytical functions to characterize various visual attributes of scenes. In general, most neural rendering methods use mesh representation to describe geometric scene properties. In addition to these methods, special techniques have also been developed to handle other image modalities using voxels [203], [204], mesh [198] and point cloud [205] data. Baek et al. [206] used the 3D mesh renderer developed in [198] to synthesize 3D hand shapes and poses from RGB images.

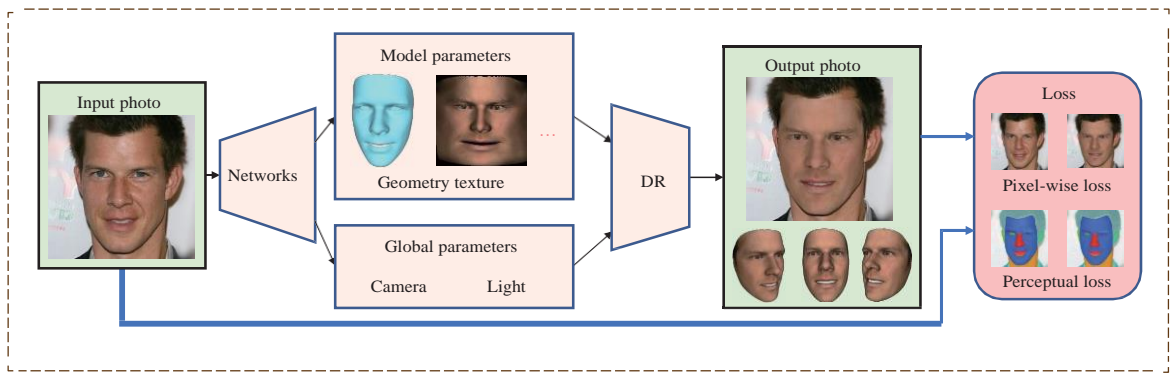

Figure 17. Simplified pipeline of the neural rendering process [197]. Reverse rendering allows intermediate 3D scenes and associated scene parameters to be generated from 2D images.

| Representation | Main strengths | Weaknesses |
|---|---|---|
| Point cloud | Low memory requirements | Low accuracy of scene topology information |
| voxel | More accurate with less processing, simplicity | High memory footprint |
| mesh | Provides more grounding (i.e., physics-aware scene representation) | High computational cost; Difficulty in describing complex shapes |
| multimodal | High resolution, more robust to visual artifacts | More complex, high computational demand |
| Implicit (NN) | Naturally differentiable, low memory requirements | Lack of grounding |
| Hybrid (Explicit + NN) | Robustness, flexibility and multipurpose | Complex architecture |

Figure 18. A summary of the main advantages and disadvantages of geometric prior representation approaches used in differential neural rendering pipelines. Here, the designation NN stands for neural network.

While approaches based on explicitly modeled geometric priors can leverage readily available 3D CAD models, they are typically limited by imperfections in the underlying models. Additional processing is usually needed to meet the requirements of specific tasks. Thus, rendering methods based on explicit representation pipelines can be adversely impacted by scene capture process and modeling time. Besides, these approaches often require the use of proprietary software tools at some stages of the development process, making accurate scene representation difficult and costly to obtain, and thereby limiting their scope of application. To achieve competitive performance, highly detailed geometric primitives and accurate scene parameters are required, along with sophisticated rendering methods. The process of modeling these elements is therefore extremely tedious. Because of these challenges, some works propose to deeply learned to compose scene primitives rather than explicitly modeling all elements from scratch

### 622    Implicit (neural) Representation

Unlike real 3D primitives whose construction is manual and laborious, their pure neural counterparts are generated automatically and can be constructed using less human effort, albeit with long training schedules. However, their ability to model 3D scene structure is directly dependent on the representation power and capacity of the underlying neural network.

To improve the effectiveness of approaches designed using scene prior representation some works [204], [207], [208] utilize deeply learned priors as auxiliary elements to refine the accuracy of explicitly modeled priors. Thies et al. [204], for instance, proposes to alleviate the burden of rigorous modeling of textures by incorporating so-called neural textures, a set of learned 2D convolutional feature maps that are obtained from intermediate layers of deep neural networks in the process of learning the scene capture (process). The learned textures are then superimposed on the geometric priors (i.e., 3D mesh) used for rendering. The approach allows course and imperfect 3D models without detailed texture information to leverage artificially generated textures to generate high-quality images. In [208], point-based neural augmentation method is proposed to enrich point cloud representations by leveraging learnable neural representations. Similarly, Liu et al. in [204] propose a hybrid geometry and appearance representation approach based on so-called Neural Sparse Voxel Fields (NSVF). The method combines explicit voxel representation with learned voxel-bounded implicit fields to encode scene geometry and appearance. While the above studies [204], [207], [208] employ learnable elements to refine scene primitives explicitly constructed from scratch, some recent works [200], [201], [202], [209], [210] have proposed to learn scene models – including geometry and appearance – entirely by using learnable elements. These representations are commonly learned using 2D image supervision (as depicted in the reverse rendering process of Figure 17). Most recent implicit neural representations commonly use Neural Radiance Fields or NeRFs [201], [209], [211], [211], an approach that employs neural networks to learn a 3D function on a limited set of 2D images to synthesize high-quality images of unobserved viewpoints and different scene conditions based on ray tracing techniques.

Currently, the photorealism of scenes generated using approaches based solely on learned representations have not matched those generated using explicit and hybrid representations. Duggal et al. [212] suggest that the problem is as a result of a lack of robust geometric priors in neural representation. They then propose an encoder sub-model to initialize shape priors in latent space. In order to guarantee that the synthesized shapes retain high-level characteristics of their real-world counterparts, high-dimensional shape priors realized with the aid of a discriminator sub-model. This serves as a regularizer of the shape optimization process. More recently, implicit neural rendering techniques have been used within generative modeling frameworks based on VAEs [213], [214] and (GANs) [48], [49], [215] to enable 3D-aware visual style transfer. These have improved the results of data synthesis achievable with either neural rendering or generative modeling techniques alone. For neural rendered scenes, the use of generative modeling helps to easily adapt visual contexts as needed. On the other hand, while generative modeling methods alone have been used to successfully model 3D scene, they often lack true 3D interpretation sufficient for complex 3D visual reasoning tasks [216].

## 6.3 Approaches and use-cases of neural rendering in data augmentation

For the purpose of data augmentation, there are three broad scenarios in which neural rendering is commonly applied:

- 2D to 2D synthesis – situations where one synthesizes additional pixel (2D) data using 2D supervision
- Scene reconstruction (2D to 3D)–3D visual understanding tasks where 3D dataset is inaccessible, and it is required to synthesize 3D objects or scenes using available 2D image data
- 3D to 2D synthesis–tasks where 2D images are generated from 3D assets

### 6.3.1 2D to 2D synthesis

A straight-forward application of the data synthesis architecture presented in Figure 17 is to generate new 2D image data with desired visual attributes from a given 2D input image. In this case, the task of the neural rendering process is to apply appropriate transformations in the intermediate

3D representation before the final rendering stage to generate desired 2D output. The basic idea is to disentangle pertinent factors of image variation such as pose, texture, color and shape. These can then be manipulated in an intermediate 3D space before mapping into a 2D space during the rendering process. Such an approach facilitates a more semantically meaningful manipulation of various scene elements and visual attributes. In data augmentation, this may be necessary when synthesizing different poses of objects in a scene [222] or when generating novel views from a single image sample (e.g., in [223]) or when introducing lighting effects to global scene appearance in order to encode invariance to these variations. These transformations are usually difficult to realize accurately with 2D operations that lack the 3D processing stage.

### 6.3.2 2D to 3D synthesis (scene reconstruction)

Another important application of differentiable rendering related to data augmentation is in scene reconstruction, i.e., the conversion from 2D image to 3D scene. Scene reconstruction techniques typically rely on inverse graphics principles. In this case, a 2D image is used to recover the underlying 3D scene. Important scene parameters such as scene geometry, lighting, camera parameters, as well as object properties such as position, shape, texture and materials are also estimated in the reconstruction process. This is useful in applications that require 3D visual understanding capabilities. Examples of such applications include for high-level cognitive machine vision tasks such as semantic scene understanding, dexterous control and autonomous navigation in unstructured environments. Tancik et al. [202] recently proposed Block-NeRF, a technique to enable the synthesis of large-scale environments (see Figure 21) . They addressed current limitations of NeRF-based models by dividing the scene representation into distinct blocks that can be rendered independently in parallel and combined to form a holistic contiguous virtual environment for training machine learning models on navigation tasks. These appearance modifications can also be applied in a blockwise manner, where smaller regions corresponding to individual NeRFs are updated separately.

Scene reconstruction methods have also been widely used to improve the quality of medical image capture in application like MIR (e.g., [224], [225] and CT (e.g., [226], [227], [228], [229], [230]). For instance, the works in [226], [227], [228] propose using implicit representations to increase the resolution of otherwise sparse CT images. Gupta et al. [229] employ NIR in an image reconstruction model, known as NeuralCT, to compensate for motion artifacts in CT images. Their approach does not require an explicit motion model to handle discrepancies resulting from patient motion during the capture process.

### 6.3.3 3D to 2D synthesis

Pixel image synthesis from 3D scene is a reverse process of scene reconstruction. It is aimed at learning the 3D scene and mapping it to a 2D space with the help of neural rendering techniques. The idea is to utilize 3D models designed by traditional graphics tools or game engines to generate 2D pixel images. The process is quite simple: given a set of primitive 3D geometric priors and corresponding scene parameters, the task is to obtain a 2D pixel image. The 3D to 2D subtask can be seen within the differential diagram in Figure 17. Unlike the 2D to 2D synthesis methods discussed earlier that require pixel images to synthesize training data, 3D to 2D synthesis methods directly generate 2D images from 3D assets.

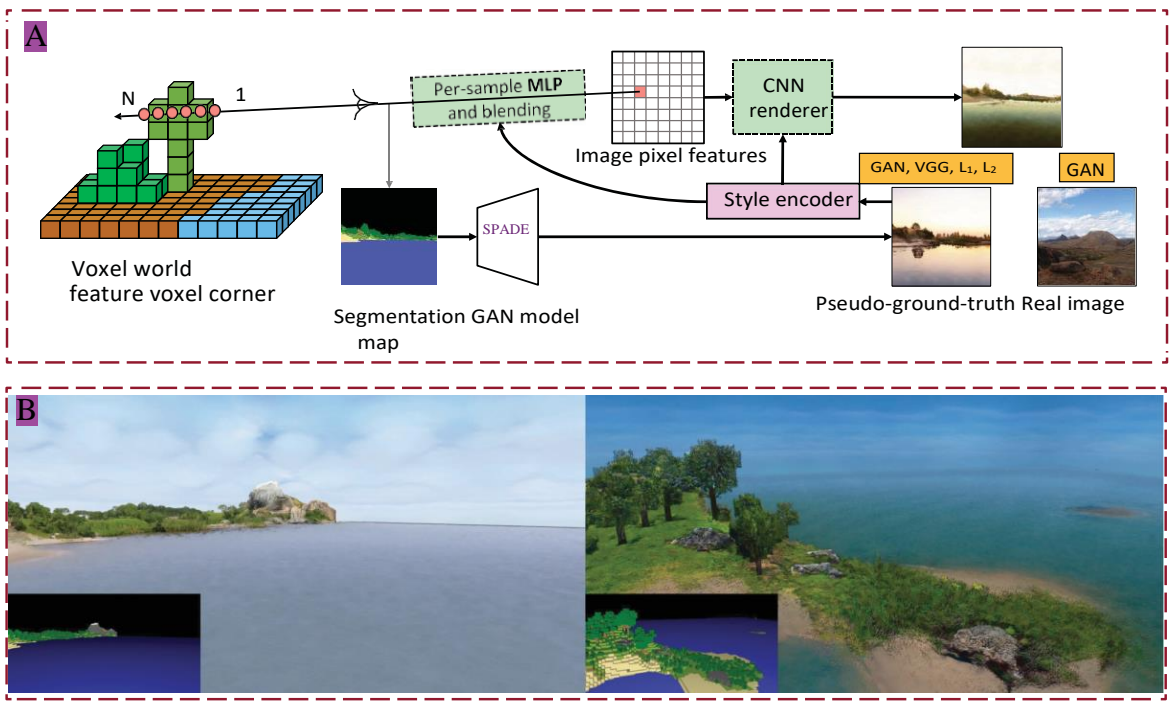

Figure 19. A: General architecture of GANcraft [217],a NeRF framework for generating realistic 3D scenes from semantically-labeled Minecraft block worlds without ground truth images. It employs SPADE, a conditional GAN framework proposed by Park et al. in [218] to synthesize pseudo-ground truth images for training the proposed NeRF model. B: Sample Minecraft block inputs (insets) and corresponding photorealistic scenes generated by the model. .

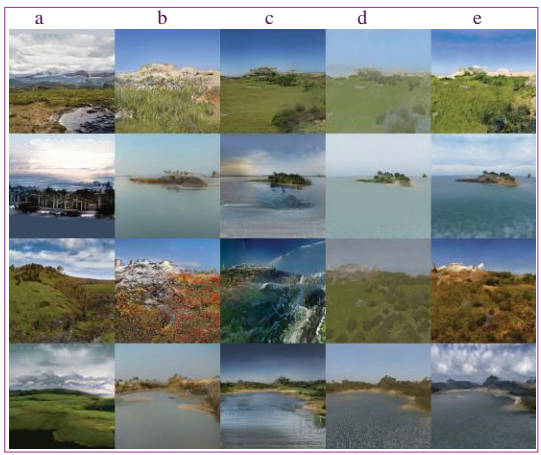

Figure 20. Visual comparison of scenes generated by different GANs and NeRF models. Each of the rows represents a unique scene, while the scenes in different columns correspond to different style-conditioning images. The specific models used here are: (a) MUNIT [219], (b) SPADE [218], (c) wc-vid2vid [220], (d) NSVF-W [221], and (e) GANcraft [217].Images are taken from [217]. .

A common application of differential rendering is to employ deep neural network models to generate 2D images and to provide flexibility in applying various 3D-like transformations in synthetically generated 2D image data. For instance, differential renderers can utilize scene elements such as 3D geometry (e.g., vertices of volumetric objects), color, lighting, materials, camera properties and motion to faithfully synthesize unobserved pixels images. Variability in the synthesized images is achieved by manipulating individual factors that contribute to the scene structure (geometry) and appearance (photometry). Thus, the process provides a way to control specific objects in the scene (e.g., modify object scale, pose or appearance) as well as general scene properties. This can help to incorporate more physically-plausible semantic features into synthesize than with traditional image synthesis approaches that lack 3D grounding.

### 6.4 Limitations of neural rendering methods and possible workarounds

Neural rendering approaches allow to bypass the tedious and laborious process of manually constructing computer graphics models of real world objects. They provide automated way for modeling complex physical processes and scenes. State-of-the-art differential neural rendering techniques make it possible to predict the spatial structure of a complex scene pixel images. They can be used to restore 3D information for a particular 2D image. Modern approaches based on NeRF models such as PixelNeRF [211], DS-NeRF [231] and IBRNet [232] can even generate diverse scenes in various lighting conditions and poses from just a few or a single RGB image. However, training such models requires enormously large quantities of task-relevant data. These approaches generalize poorly to unseen data. Since, in many practical settings, labeled datasets for these tasks may be extremely challenging to produce, many works [207], [208], [233], [234] incorporate traditional graphics pipelines into end-to-end deep neural networks to view-consistent scenes. Generative modeling techniques [47], [215], [217] have also been suggested as a viable means for providing the necessary supervision in the absence of ground-truth images. Hao et al. in [217], for instance, propose a neural rendering model that allows to generate realistic scenes from simple 3D LEGO-like block worlds like Minecraft. Since there are no ground-truth images for these types of inputs that can be used to supervise the training, the authors utilize a pretrained GAN-generated images as pseudo-ground truth instead of real images. Their approach, like [202], also allows flexible user control over both scene appearance and semantic content using appearance codes. This capability is useful for applications like long term visual tracking [235], where scene dynamics impacts severely on performance; the ability to control scene appearance is extremely important to simulate all possible view conditions. The basic structure of GANcraft and sample outputs are shown in Figure 19. In Figure 20 we present a visual comparison of scenes generated by different GANs (MUNIT [219], (SPADE [218] and wc-vid2vid [220]), NeRF (NSVF-W [221]) and a combination of GAN and NeRF (GANcraft [217]).

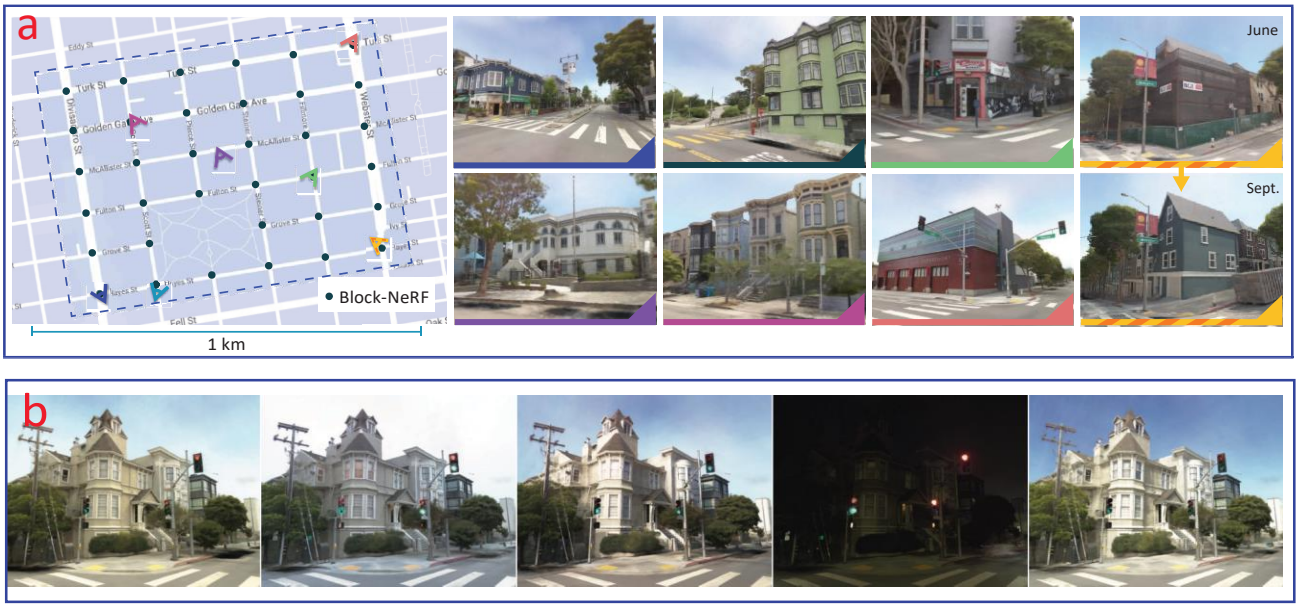

Figure 21. State-of-the-art neural rendering models such as Block-NeRF [202] allow large 3D environments to be generated from sparse 2D views. In multiple neural radiance field models combine to encode large. Shown here is an approximately one square kilometer environment rendered by Block-NeRF (a). The authors also provide a way to alter the appearance of the rendered scene corresponding different environmental conditions such as weather and illumination changes (b). .

An important aspect of data synthesis based on neural rendering approaches is the ability to work with different data representations – for example, voxels, raw point clouds, pixels, or implicitly defined data forms based on learned functional description of the physical properties of objects. However, this flexibility also presents difficulties: representations can be very different for the same data and the different forms of representations are not necessarily compatible with one another. Recent techniques [236], [237], [238], [239] allow to fuse or convert between these diverse data representations. However, these methods often lead to the loss of vital information about the scene elements. As a result of these difficulties, training neural rendering models to generate data in these modalities is still a challenging issue. Moreover, relatively small number of datasets exist for training differential neural rendering models on many of the possible representation modalities. Given the rapid growth of interests in neural rendering methods, this problem will be solved in the short term.

Implicit neural representation can also support non-standard imaging modalities such as synthetic aperture sonar (SAS) images [240], [241], computed tomography (CT) [230], [242], [243] and intensity diffraction tomography (IDT) [244]. In addition, these approaches are capable of modeling non-visual information like audio signals [245]. Approaches have also been proposed to leverage multiple visual modalities together with other physical signals to provide complementary information about the physical properties of objects [245], [246], [247]. This can support multi-sensory intelligence in applications such as robotics, virtual, augmented, extended and mixed reality, and human-computer interaction. However, adding these additional elements means that data requirements grow even more exponentially, making it challenging to accomplish realistic results based on only learnt models. Many works (e.g., [248], [249], [250]) propose integrating task-specific priors to improve the ability of neural rendering methods that model complex scenes and interactions to generalize better to unseen contexts. Incorporating strong priors often requires special mathematical formulations within neural network models for encoding spatial features and natural behavior of objects in the 3D world, leading to increased computational complexity and cost. For this reasons, state-of-the-art neural rendering frameworks such as [248], [250], [251] are inherently complex, limiting their ability to represent large-scale scenes. Block-NeRF [202] proposes a modular framework that allows separate NeRF modules to represent individual regions of a scene (shown in Figure 21). This allows large-scale 3D scenes to be represented and efficiently manipulated using sparse 2D images.

# 7 NEURAL STYLE TRANSFER

Neural style transfer (NST) is a method for synthesizing novel images similar to GAN-based style transfer. However, in contrast with generative modeling approaches, neural style transfer exploit conventional feed forward convolutional neural networks for the synthesis.

## 7.1 Principles of neural style transfer

Neural style transfer involves first learning representations for the content and structure of the original images, and the style of a reference samples. These representations are then combined to generate new representations in the style of the reference images while at the same time maintaining the content and structure of the original image. The method leverages the hierarchical representation mechanism of deep convolutional neural networks (DCNNs) to flexibly generate novel images with various appearance artifacts and styles. An illustration of the basic principles of the concept is shown in Figure 22. Since shallower layers in CNNs encode low-level visual features such as object texture, lines and edges [252], while deeper layers learn high-level semantic attributes, different augmentation schemes can be realized by manipulating the two semantic levels separately and combining them in different ways. In NST, typically, a DCNN model without the fully connected layers are used to extract image features at different levels. Low level features encoded by shallower layers are then extracted and combined with high-level features extracted from a second image. As the second image contributes high-level features, essentially, its semantic content is transferred to the artificially generated image while the first image's visual style is transferred (see Figures 22 and 23).

The original technique for neural style transfer) was first proposed by Gatys et al. in [254] as a way to artificially create different artistic styles in images. Specifically, they altered landscape images taken by digital camera to look like images produced by artworks while still maintaining their semantic meaning. To accomplish this, Gatys et al. [254] remove the fully connected layers of a VGG model [255] and use the convolutional layers to extract visual features at different semantic levels. They then introduce an optimization function that consists of two different loss components – the style and content loss functions – which are controlled separately by different sets of weight parameters. Based on these principles, the authors showed that image content and style information can be transferred independently to different visual contexts by separating the filter responses of shallower and deeper layers of the CNN model. To create a new image with the content of a source image $I_c$ and the style of a target image $I_s$, they propose to combine the latter processing stages of $I_c$ with the earlier stages of $I_s$ (Figure 22). Neural style transfer has also been extended to video [?], [256], [257], [258] and 3D computer vision [259] applications. Ruder et al. [259] propose an video style transfer technique that applies a reference style from a single image to an entire set of frames in a video sequence. In [260], a neural style transfer approach is combined with a GAN model to generate diverse 3D images from 2D views. The generated image data is then used to improve performance 3D recognition tasks.

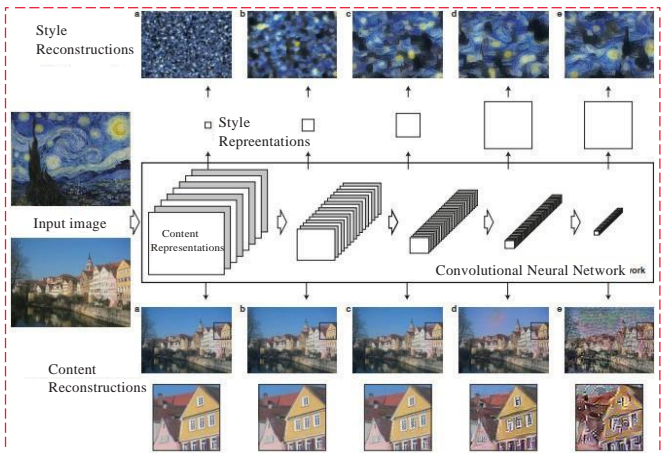

Figure 22. Neural style transfer relies on the hierarchical representation of features in convolutional neural networks to render high-level semantic content of images in different visual styles. In the original work [253] Gatys propose to extract style and content information from different processing stages of convolutional neural networks. The extracted style and content information are then manipulated separately and combined to form new images.

## 7.2 Neural style transfer as a data augmentation technique

Following the original work [254], many subsequent studies (e.g., [260], [261], [262], [263], [264], [265], [266], [267], [268]) have employed neural style transfer technique as a form of data augmentation strategy to synthesize novel images to extend training data. In many of the works, including the original work that introduced the approach, style transfer is typically applied to synthesize non-photorealistic images. It is reasonable to assume that adding images that are stylized in a non-photorealistic way to the training dataset could still help to reduce overfitting and improve generalization performance. Indeed, a number works have shown this to be the case for many tasks. Jackson et al. [269] demonstrated that augmenting training data nonphotorealistic images (e.g., images synthesized in artistic styles)can significantly improve performance in several different computer vision tasks. Using neural style transfer, Jackson et al. in [269] achieved an improvement in the range of 11.8 to 41.4% over a baseline model (InceptionV3 without augmentation) and improvement between 5.1 and 16.2% over color jitter (alone) on cross-domain image classification tasks. In addition, their method yields a performance increase of at least 1.4% over a combination of seven different classical augmentation types. Instead of transferring a single style per image, the authors in [269] aim to create more random styles suitable for multi-domain classification tasks by randomizing image features (texture, color and contrast). They used a so-called style transfer network to obtain random style attributes by sampling from multivariate distribution of low-level style embeddings.

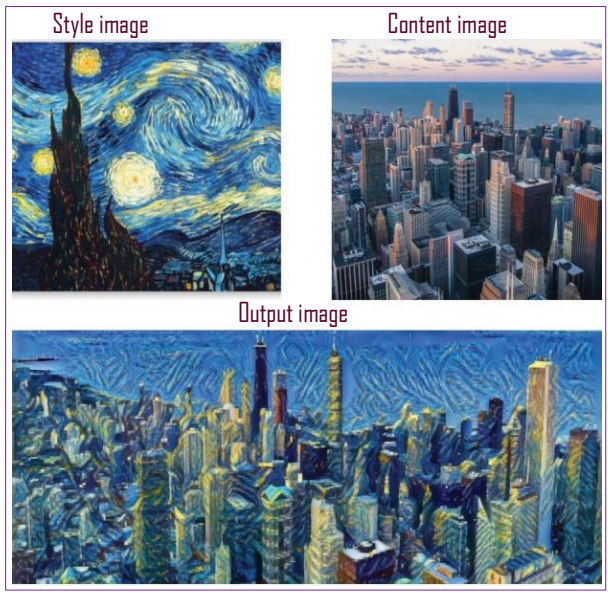

Figure 23. An example of artistic style transfer by neural style transfer (NST) technique (image courtesy [270]). The NST model reproduces the content of the content image in the style of the style image.

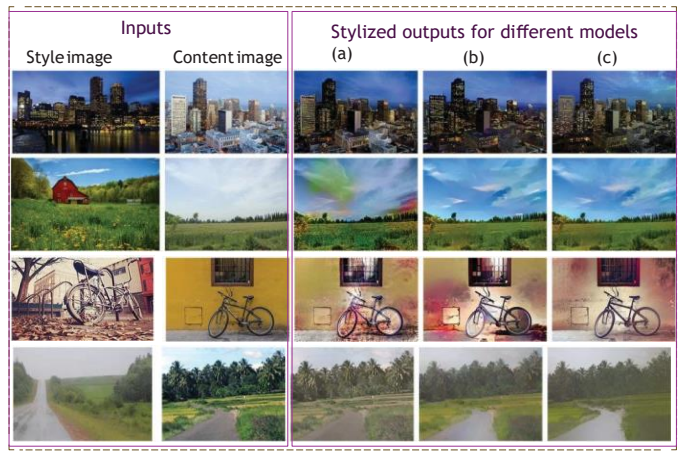

Figure 25. Visual comparison of different NST-based photorealistic stylization methods. We show results for Li et al. [274] (column a), Luan et al. [275] (column b) and Yang [276] (column c).

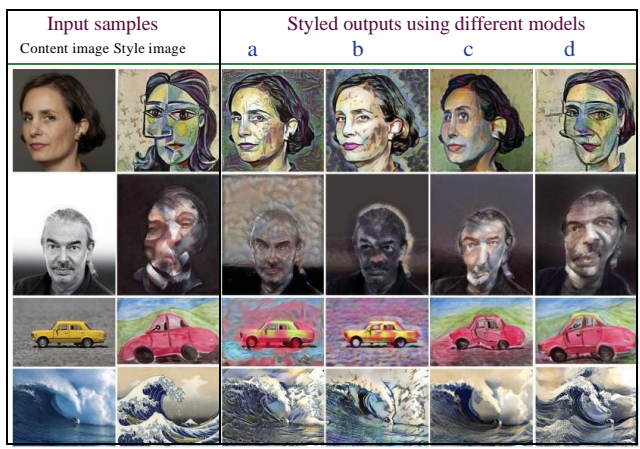

Figure 24. Comparison of results of different artistic style transfer approaches: Gatys et al. [253] (a), Huang and Belongie [271] (b), Kim et al. [272] (c) and Liu et al. [273] (d).

## 7.3 Approaches to data augmentation using NST

In the literature, three main approaches to data augmentation based on neural style transfer can be distinguished. These are (1) approaches aimed at increasing data variability in order to improve generalization performance, (2) approaches aimed at synthesizing photorealistic image data, and (3) approaches that seek to selectively modify only the most important (in a given context) image parts. We briefly discuss these three neural style transfer implementations in subsections 7.3.1 to 7.3.3. In Table 3, we present a summary of the main features of these approaches.

### 7.3.1 Inproving data diversity with artistic stylization

Since higher data diversity generally favor better generalization accuracy, many works [281], [285], [286], [287] have recognized the importance of increasing diversity of styled images and have introduced different techniques to achieve this. Wang et al. [281] propose an approach that relies on the perturbation of deep convolutional features by means of random noise injection. Another class of approaches [263], [280], [288], [289] aim to address this problem by developing methods that allow many different styles to be transferred simultaneously. For instance, in [280], sets of convolution feature maps in intermediate CNN layers are grouped into filter banks, with each group uniquely encoding one specific style (see Figure 27). Dumoulin et al. [289] propose a simple normalization approach – named conditional instance normalization – that leverages the mean-variance statistics of convolutional features to enable as many as 32 different styles to be transferred to a specified content image. Instead of explicitly extracting style information from reference images, Georgievski [263] proposes an approach in which different styles can be independently specified as input style embeddings. First, a large set of generic style embeddings are first obtained from a pre-trained neural network (Inception-ResNet-V2) that is trained on large-scale image dataset (in this case, ImageNet). The learned styles are then used to construct a multivariate normal distribution of styles that can later be applied to content images. The application of styles is accomplished through an encoder-decoder sub-network, which the author termed style transformer. Huang and Belongie [271] extend the concept of [289] by modifying the instance normalization operation in such a way that it is able to encode arbitrary style, as opposed to transferring a pre-defined set of style as in [289]. They introduced a new layer, adaptive instance normalisation (AdaIN), in place the conditional instance normalization layer in [CIN]) whose function is to align the mean and variance of convolutional features of style and content images. Gu et al. [288] propose to reshuffle deep feature maps of the style image so as to synthesize images with arbitrary styles. Data augmentation schemes based on such stylizations would result in more ro-bust representations akin to style randomization in rendered images.

Table 3
The main ways to perform data augmentation using neural style transfer technique

| Approach | Rationale for data augmentation | Sample works |
|---|---|---|
| Artistic style transfer | Simulates noise to provide variability in training data so as to decouple models from being tied to specific visual features. It can also help encode invariance to image textures. | [261], [263], [269] |
| Photorealistic style transfer | Simulates desired real-world visual effects such as illumination and weather effects in training data. | [277], [278], [279] |
| Multi-style transfer | Increase the diversity of training images to improve generalization performance | [280], [281] |
| Patch-level stylization | Provides a means to manipulate specific image content (e.g., objects of interest in object detection task) | [267], [282], [283] |

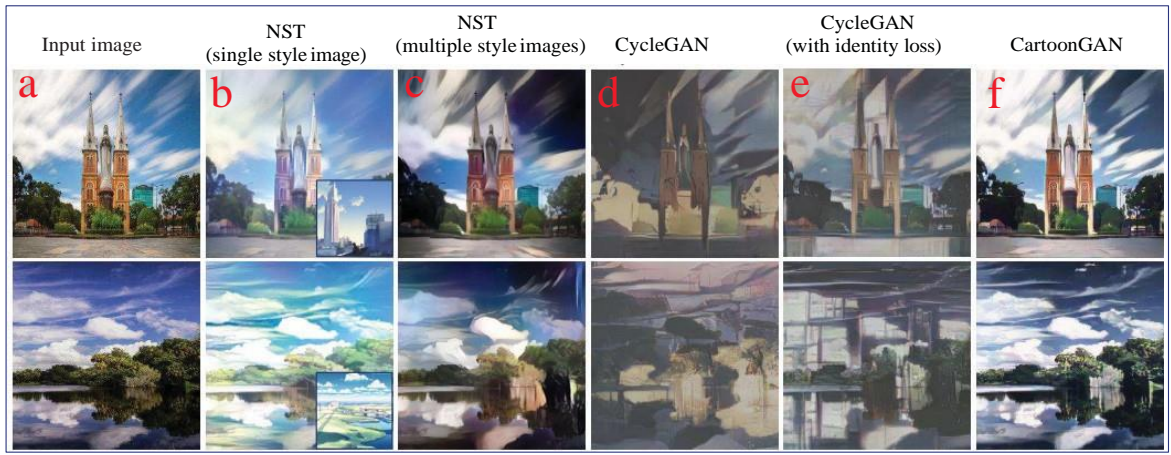

Figure 26. Qualitative comparison of neural style transfer (NST) and GAN models (results courtesy [284]). The different models aim to perform artistic stylization of the original images while maintaining all content. The styles were obtained from images extracted from cartoon videos. The NST model in column b has been trained on style image that has similar content as the input image (shown inset). The rest of the columns (c, d, e and f) show results of styling with an aggregate of 4,573 cartoon images. Note that the quality of stylization is defined by how well the semantic content is preserved, including the clarity of edges. The specific models used are: NST by Gatys et al. [253] (column b and c), CycleGAN [45] (column d), CycleGAN with identity loss (column e), and CartoonGAN [284] (column f).

In addition to these approaches that aim to improve image synthesis by introducing new architectural methods of encoding styles within CNN layers, a number of works [290], [291] are primarily focused on developing new loss functions that allows to better transfer more diverse and fine-grained features than earlier approaches based on style and content loss functions [254]. Li et al. in [285] introduced a special loss function, known as diversity loss, to encourage diversity. Wang et al. Luan et al. [275] propose a loss function based on smoothness estimation.

### 7.3.2 Photorealistic data syntheis with NST

Recently, several researchers [275], [292], [292] have introduced methods to enhance the photorealism of images synthesized using NST methods. Photorealistic styles are stylization schemes that render the resulting image as visually close as possible to real images captured by real image sensors. Photorealism is aimed at simulating high-quality data in terms of image details and textures, taking into consideration a range of real-world factors, which may include blurring effects resulting from camera motion, random noise, distortions and varying lighting conditions. Figure 24 compares the results of different artistic style transfer approaches. In Figure 25, the outputs of various models that aim to achieve photorealistic stylization are compared. We also compare the artistic style transfer ability of common GAN models with NST methods in Figure 26.

There is also a family of NST approaches that aim to transfer specific visual attributes such as image color [286], [293], [294], illumination settings [295], material properties [296] or texture [297], [298]. Rodriguez-Pardo and Garces in [296] propose a NST-based data augmentation scheme based on transferring properties of images under varying conditions of illumination and geometric image distortions. Gatys et al. in [299] propose to decompose style into different perceptual factors such as geometry, color and illumination which can then be combined in different permutations to synthesize image with specific, desirable attributes. In order to avoid geometric distortions and preserve statistical properties of convolutional features extracted from style images, Yoo et al. [292] introduced a wavelet corrected transfer mechanism that replaces standard pooling operations with wavelet pooling units.

Neural style transfer techniques have also been successfully employed to transfer specific semantic visual contexts to synthetic data. For instance, Li et al. [277] transferred images taken in clear weather to snowy conditions. In [253], images taken in bright day are converted to night images using NST stylization techniques. More advanced neural stylization approaches [278], [279] have been designed to learn the semantic appearance as well as physically-plausible behavior of objects of interest. Kim et al. [278], for instance, proposed a physics-based Neural Style Transfer method to simulate particle and fluid behavior in synthetic images. Their approach allows realistic visual appearance and semantic content of different particular matter to be learned and transferred from 2D to 3D settings.

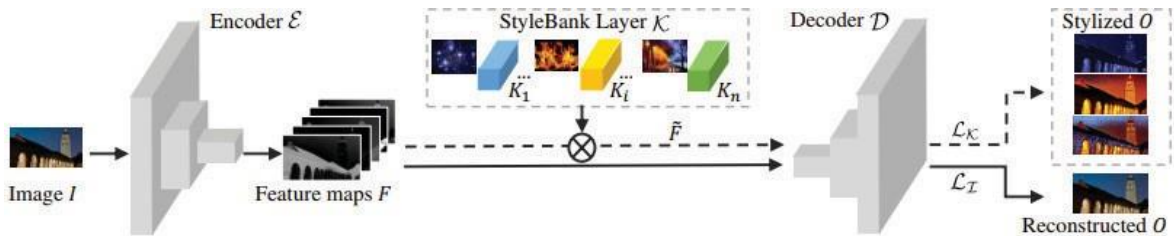

Figure 27. StyleBank [280] is an example of multi-style transfer technique based on an encoder-decoder architecture. It encodes different styles in groups of convolutional filters in intermediate CNN layers that can be selectively applied to specific image content.

### 7.3.3 Patch-level synthesis

Researchers have also proposed photorealistic synthesis methods [296] based on patch-wise stylization that allow specific image parts or object within images to be selectively altered. Such methods allow a more granular control of the output images. They may enable, for instance, the manipulation of specific objects in a complex scene. This family of style transfer techniques can be useful in applications such as semantic segmentation (e.g., in [282]. Some works have suggested using patch-level styling to achieve photorealism while at the same time reducing the computational demands of conventional (holistic image) methods. Cygert and Czyżewski [267] argue that styling the whole input image could be detrimental to generalization performance. They propose to address this limitation by utilizing a stylization method that transforms only small patches of input images. To achieve this, they perform a hyperparameter search to find the best patch size for stylization. In [283], Chen et al. proposed a patch-based stylization technique as a means of reducing complexity and improving memory efficiency in the stylization of high resolution images.

### 7.4 Limitations of NST methods and possible workarounds

The main advantage of neural style transfer is the fact that the approach leverages the natural representation of visual features in CNN layers to synthesize new image data. Consequently, the amount of data and the time required for training models is relatively small. Also, model architectures for neural style transfer are generally simpler compared to other synthesis methods such as generative modeling and neural rendering. Moreover, since NST training is usually more sample-efficient than other synthesis methods, memory requirement is greatly reduced.

One major weakness of datasets created by neural style transfer approaches is that the data is generated by feature-level manipulations and not by intuitive transformation of real data. As a result, it is extremely difficult to account for semantic information about the entities in a scene when applying stylization. This can also lead to noise and unwanted artifacts which may seriously affect the quality of the generated data. Another challenge with non-intuitive, feature level perturbations is the difficulty in maintaining consistency between local and global features [275] when applying styles. These issues require additional (costly) processing of semantic information so as to ensure that realistic data is produced. For instance, recent works have proposed to optionally perform semantic segmentation on the target scene so as to enable context-specific style transfer (e.g., in [275], [300]). This additional processing step may lead to increased model complexity and computational cost. In addition, it can potentially introduce errors, e.g., a result of incorrect segmentation masks, which may harm the accuracy of the stylization.

Also, the fact that the generated dataset is not derived from intuitive, physically grounded manipulation of input data at the sample level may limit the degree to which desired visual attributes can be simulated. Unlike neural rendering and generative modeling techniques that can naturally learn desired visual contexts such as color, contrast and illumination levels, neural style transfer methods require more complex formulations to handle these attributes successfully.

An even more serious limitation of neural style transfer methods is the difficulty of simulating spatial transformations. For this reason, the approach is mainly used for photometric data augmentation tasks. Furthermore, geometry-consistent photometric stylizations (e.g., brightness levels based on viewpoints; generating and placing shadows at the right positions in a scene; and fine-grained style-content consistency under different poses) are difficult to achieve. This can lead to distorted style patterns. The difficulty lies in the inherent principle of the approach: different hierarchical levels of feature maps separately capture style and geometry information and are combined to generate new samples. The decoupling of style from geometry makes it challenging to apply learned styles in a view-dependent manner. Some works (e.g., [273], [301]) have proposed to incorporate explicit spatial deformation models within NST architectures to handle geometric transformations. In [302] Jing et al. utilize graph neural network model to learn fine-grained style-content correspondences that minimizes local style distortions under geometric transformations.

## 8 SYNTHETIC DATASETS

Presently, a wide range of large-scale synthetic datasets are publicly available for training and evaluating machine vision models. We summarized the details of some of the most important synthetic datasets in Table 4. These datasets support a wide range of visual recognition tasks. In addition, they cover many of the synthesis methods explored in this work. In addition, they include the common representation methods.

Table 4
A summary of publicly-available synthetic datasets

| Dataset | Synthesis method | Domain | Supported tasks | Dataset size |
|---------|------------------|--------|-----------------|--------------|
| RTMV [303] | Neural rendering | General scene understanding | View Synthesis<br>Scene reconstruction<br>Pose estimation | 300,000 |
| MPI-Sintel [161] | 3D animation (video) | Spatio-temporal scene understanding | Optical flow | 1628 |
| Objectron [304] | 3D data from augmented reality library | Multi-view object recognition | 3D object detection | 4M |
| Synthia [164] | 3D game engine | Autonomous driving | Depth estimation<br>Scene segmentation<br>Ego-motion | 200,000 |
| LCrowd [166] | 3D game engine | Crowd analysis | Crowd counting<br>Person detection (in crowd) | 20M |
| Semantic3D [305] | Point cloud data from laser scanner | urban scene understanding | Semantic segmentation | |
| Synscapes [165] | Neural rendering | Autonomous driving | Pose estimation<br>Depth estimation<br>Semantic and instance segmentation | 25,000 |
| ShapeNet [146] | 3D CAD | General recognition | 3D Perception | 51,300 |
| ScanNet [172] | RGB-D capture | Indoor scene understanding | Semantic and instance segmentation | 2.5M |
| Virtual Kitti [162] | 3D game engine | Urban scene understanding | Depth estimation<br>Optical flow<br>Object detection and tracking<br>Scene and instance segmentation | 21,260 |
| Pix3D [306] | 3D scanning and web sources | Object shape retrieval viewpoint estimation | 2D-to-3D reconstruction | 395 |
| GTA-V [160] | Video game play | Autonomous driving | Sematic Segmentation | 25,000 |
| Hypersim [143] | 3D CAD | Indoor scene understanding | Semantic and instance segmentation | 77,400 |
| PreSIL [307] | Video game play | 3D scene understanding | 2D and 3D object detection<br>Depth perception<br>Semantic segmentation | 50,000 |
| SOMASet [148] | 3D CAD | Person re-identification | Object recognition | 100,000 |
| SceneNet-RGBD [308] | RGB-D rendered objects from CAD models | Indoor scene understanding | Object detection<br>Pose estimation<br>Depth estimation<br>segmentation | 5M |

# 9 Effectiveness of synthetic data augmentation methods

Many works have demonstrated the effectiveness of synthetic data augmentation techniques. In some cases (e.g., [3], [4], [184], [309]) data generated synthetically leads to better generalization performance than real data. Wang et al. [309], for example, reported several instances where models trained on synthetic data achieve better results on face recognition tasks. Similarly, Rogez and Schmid [4] consistently obtained higher performance with synthetic data than with real data on pose estimation tasks. Applications scenarios where synthetic images have particularly performed better than real data have been settings that do not require high level of photorealism (e.g., depth perception [3]) and pose estimation [4], [184]. This is due to the fact that synthetic images are often "cleaner" (i.e., the do not contain spurious details and artefacts which may be irrelevant to the target task). While some studies have obtained impressive results with training exclusively on synthetic data, many other works show that synthetic data, when used alone, would not always yield the desired performance. For instance, results obtained by Richter et al. in [160] demonstrates that while synthetic data can drastically reduce the amount of real training data needed to achieve optimum performance, by themselves synthetic images do not guarantee good performance. In their study [160], they experimented first with real data and obtained an mean IoU of 65.0%. When the training set was augmented with synthetic data, they were able to improve the mean IoU score by 3.9% (from 65.0 to 68.9%). Indeed, using the augmented data, they achieved comparable performance (65.2%) as with real data (65.0%) using only about one-third of the real data. However, the results were poor and unsatisfactory when only synthetic data was used (43.6%). Rajpura and Bojinov [310] compared the performance of deep learning-based object detectors trained on synthetic (3D-grahics models), real (RGB images), and hybrid (synthetic and real) data. The results showed lower performance on synthetic data (24 mAP) compared to real (28 mAP). However, the addition of the real and synthetic images improves the performance by up to 12% (36 mAP). Similarly, in [177] Alhajia et al. observed that training with augmented reality environment that integrates real and synthetically generated objects into a single environment achieves a significantly higher performance than with either separately. In [311], Zhang et al. observed that expanding the training set by increasing the proportion of synthetic data does not lead to a linear increase in model performance. In fact, for some tasks, the performance flattens out at about 25% composition of synthetic data for the specific cases investigated.

## 10 SUMMARY AND DISCUSSION

In the literature, many data synthesis approaches have been proposed. Despite the large variety of techniques, four main classes of approaches can be distinguished: generative modeling, data synthesis by means of computer graphics tools, neural rendering approaches that utilize deep learning models to simulate 3D modeling process, and neural style transfer that relies on combining different hierarchical levels of convolutional features to synthesizer new image data. The first group of approaches – generative modeling – is mainly based on the generative adversarial networks and variational autoencoders. Generative modeling methods allow realistic image data to be synthesized using only random noise as input. Models can also generate outputs conditioned on specific input characteristics that define the desired appearance of target data. The second class of methods, computer graphics approaches explicitly construct 3D models based on manually modeled, primitive geometric elements. Neural rendering methods use deep neural networks to learn the representation of 3D objects and then optionally render these as 2D images. Neural style transfer combines different semantic information contained in different layers of neural networks to synthesis various visual styles. The main strengths and weaknesses of each of these classes of approaches are summarized in Figure 28.

The different data synthesis approaches have unique characteristics that define their scope of application. Neural style transfer method, for example, is highly flexible since the stylization process can be controlled by setting appropriate weight parameters corresponding to different style intensities. This allows low-level transformations to be easily accomplished for a given task. However, despite its simplicity and relative efficiency, the method is generally limited to 2D synthesis. It is also challenging to generate large-scale photorealistic data. Its most important prospect is in enhancing robustness to noise, overcoming overfitting by preventing models from learning specific visual patterns, and encouraging texture invariance. Unlike neural style transfer methods, rapid advances in generative modeling techniques have made it possible to easily create large volumes of artificial images with desired properties for specific tasks. However, these methods typically model objects and scenes in 2D, making it difficult to transfer encoded knowledge to 3D scene understanding tasks. Also, by ignoring the 3D structure of the real world, achieving physically-grounded object manipulation is a challenging task. Synthesis methods based on 3D graphics modeling overcomes these limitations by providing a means to generate realistic 3D scenes. However, the modeling process is typically laborious, limiting the amount of detail and the range of dynamic attributes that can be supported. Neural rendering can be used to introduce more nuanced details without manually creating the desired visual appearance. In particular, differential neural rendering approaches based on implicit representations allow deep learning models to encode 3D-grounded representation, as well as visual attributes such as color and varying lighting conditions in the artificial generated images. Only in the last few years have there been cost-effective methods for generating large-scale virtual scenes using implicit neural representation. Recently, NeRF been a subject of considerable research. This intensive research has led to the development of new deep learning methods and models that adequately represent 3D data in sophisticated and efficient ways. As a consequence of the development these methods, a wide range of possibilities are emerging regarding the training data for complex tasks like autonomous driving, where the cost of camera-captured data is exorbitant. As well as being able to generate more plausible pixel images, modern differential neural rendering models such as Block-NeRF [202] and Mega-NeRF [312] can synthesize realistic, large-scale 3D videos of entire scenes using only a few 2D images as input. These models basically map pixel space into the context of a continuous 3D scene. This capability is highly promising in complex visual perception tasks that require physics-aware interpretation of input data. Conceivably, in the near future more powerful and efficient NeRF models capable of generating large-scale, fully dynamic and realistic 4D scenes will replace the existing methods for synthesizing training data for applications such as robot navigation, autonomous driving and many 3D perception tasks.

## 11 FUTURE RESEARCH DIRECTIONS

The four main classes of synthetic data augmentation methods surveyed in this work are computer graphics modeling, neural rendering, generative modeling and neural style transfer. Overall, judging by recent trends, we expect significant progress in generative modeling and neural rendering techniques and minimal progress in neural style transfer methods. We also expect the applications of these methods in more challenging machine intelligence tasks such as affordance learning, human-machine interaction and extended reality. Progress in this areas will undoubtedly have a profound impact on the development of machine vision and artificial intelligence as a whole. We outline most promising future research directions in the following paragraphs.

**a. Modeling multiple sensory modalities in an integrated and adaptive way**

Recent interest in generative modeling and neural rendering approaches has led to the development of new hybrid deep learning methods that combine the power of state-of-the-art neural rendering techniques like NeRFs and generative models such as GANs and VAEs to adequately represent both 2D and 3D data in more effective and controllable ways. These approaches primarily exploit conditional GAN and conditional VAE techniques to provide a form of control over the visual properties of the synthesized data, allowing models to learn multiple salient features to enhance the realism of representations. This additional dimension of flexibility and control can be leveraged by future synthesis models to specifically generate more dynamic data whose appearance and properties change adaptively with the visual context. For applications such as high-level perception and long-term scene understanding, the emergence of techniques that allow view-specific attributes to be modified online in response to real-world conditions would be extremely useful. More generally, the development of new data synthesis methods based on implicit neural rendering and generative modeling techniques will offer opportunities to radically extend the capabilities of state-of-the-art machine vision systems. Because implicit neural representation techniques are spatio-temporally continuous and differentiable, in addition to 3D visual information, they can be used as universal function approximators to model diverse sensory signals as well as complex physical processes of the real world. Future research is expected to produce new and improved ways to enable these diverse sensory modalities and physical properties of objects to be compactly integrated into a kind of universal framework. Incorporating information-rich representation in this manner will undoubtedly help to improve "common sense" reasoning capabilities of intelligent and robotic systems.

| Approach | Main advantages | Disadvantages |
|---|---|---|
| Generative modeling | • Automates the process of generating synthetic data<br>• Can generate highly photorealistic data without additional effort | • Lack of 3D interpretation<br>• Requires representative examples of the target class<br>• Problems with mode collapse and overfitting can lead to poor results |
| Computer graphics modeling | • Requires no training data whatsoever<br>• End product fully controllable by developer<br>• Can support physically-plausible behaviors<br>• Can generate very large virtual environments | • Subject to bias and perceptual limitations of human developers<br>• Extremely laborious<br>• Requires extensive domain knowledge<br>• May require the use of proprietary tools |
| Neural rendering | • Can easily support 2D and 3D tasks<br>• Allows easy manipulation of generated data<br>• Can extend the capabilities of existing 3D models | • Highly complex model architectures<br>• Scene generation capacity limited to small environments<br>• Requires enormous amounts of data |
| Neural style transfer | • Relatively simple architectures<br>• Low resource requirements<br>• Easier to implement | • Limited in the ability to generate photorealistic data<br>• Can only synthesize 2D images |

Figure 28. A comparison of different data synthesis approaches.

### b. Towards more effective and efficient representation and training

A major shortcoming with current data synthesis approaches, particularly generative modeling and neural rendering techniques, is the need to use several examples from the target domain to guide the synthesis process, especially when training for 3D scene understanding tasks. Many recent works have focused on achieving more computationally-efficient representation that enables scene data to scale up to city or metropolitan level environments. Application domains such as outdoor scene understanding and autonomous driving particularly require very large continuous scenes which are currently challenging to synthesize owing to the enormous computational resource requirements. Today's large-scale synthetic environments are typically realized by stacking several smaller, image-level scenes together, where each constituent "mini scene" is encoded by a dedicated NeRF model. Obviously, this is not the most natural and effective way to represent scenes. A large amount of future research efforts will increasingly focus on achieving more sample-efficient training. One of the most promising research goals is the development of unconditional 3D-aware data synthesis techniques that allow to generate high-quality, realistic 3D synthetic scenes without the need for reference images. New synthesis methods will also allow to utilize more parse representations to synthesize realistic data with detailed visual information. The integration of neural radiance fields (NeRF) models into state-of-the-art generative modeling architectures will increasingly provide better ways to achieve more compact and unified differentiable representations of complex scenes. New techniques resulting from further breakthroughs in this line of works, in the not-so-distant future, can be used to generate seemingly continuous and infinitely large scenes by exploiting more effective and more efficient representation techniques.

### c. Towards synthesis and representation of context-relevant scene properties

Machine vision models mainly rely on visual appearance of input data to make predictions. The realism of visual features is, thus, the sole concern of developers and researchers when designing methods for synthesizing training data. However, in more complex tasks such as affordance learning, robot perception and dexterous manipulation, in addition to synthesizing appearance and geometry, it is often helpful to model non-visual properties such as friction, mass and other semantic information about objects in the scene. We expect that approaches that rely on jointly modeling visual information and high-level non-visual attributes that reflect properties of the real world – or physics-aware data synthesis – will become an important research topic in the foreseeable future. Research in this direction will facilitate new ways to represent visual and semantic context information in more unified and coherent fashion. This will allow the synthesis fully interactive 3D environments without explicitly modeling object properties and behavior to become possible.

### d. Simulating less intuitive augmentation schemes

Current data synthesis approaches assume visual similarity of the original data or target domain and the generated image samples. Consequently, the generative modeling process primarily aims to generate clean data that is as close as possible to the target data. It is, however, known that techniques such as random image perturbations can sometimes provide the most useful augmentations for improving generalization. Thus, by focusing on more aligned semantic content, data synthesis approaches based on generative modeling usually ignore useful augmentation strategies that rely on non-realistic data (e.g., methods such as blurring and noise injection). At present, there is no workaround that allows to generate implausible but effective data using generative modeling.

## 12 CONCLUSION

Synthetic data augmentation is a way to overcome data scarcity in practical machine learning applications by creating artificial samples from scratch. This survey explores the most important approaches to generating synthetic data for training computer vision models. We present a detailed coverage of the methods, unique properties, application scenarios, as well as the important limitations of data synthesis methods for extending training data. We also summarized the main features, generation methods, supported tasks and application domains of common publicly available, large-scale synthetic datasets. Lastly, we investigate the effectiveness of data synthesis approaches to data augmentation. The survey shows that synthetic data augmentation methods provides an effective means to obtain good generalization performance in situations where it is difficult to access real data for training. Moreover, for tasks such as optical flow, depth estimation and visual odometry, where photorealism plays no role in inference, training with synthetic data sometimes yield better performance than with real data.

### ACKNOWLEDGMENTS

The authors would like to thank...